\documentclass[journal,twoside, final]{IEEEtran}
\usepackage{epsfig}
\usepackage{graphicx}
\usepackage{caption}
\usepackage{amsmath}
\usepackage{amssymb}
\usepackage{float}
\usepackage{subfigure}
\usepackage{color}
\usepackage{mhchem}
\usepackage{booktabs}
\usepackage{multirow}
\usepackage{tablefootnote}
\usepackage{booktabs}
\usepackage{colortbl}
\usepackage{hyperref}
\usepackage{bbding}
\usepackage{framed}
\usepackage{algorithm}  
\usepackage[table]{xcolor}
\usepackage{algpseudocode}  

\ifCLASSINFOpdf
\else
\fi
\begin{document}
	%
	\title{Weakly Supervised Semantic Segmentation via Progressive Patch Learning}
	%
	%
	%
	
	\author{Jinlong~Li$^\dagger$,
		Zequn~Jie$^\dagger$,
		Xu~Wang,~\IEEEmembership{Member,~IEEE,}
		Yu~Zhou,~\IEEEmembership{Member,~IEEE,}
		Xiaolin~Wei,
		Lin~Ma,~\IEEEmembership{Member,~IEEE}
		\thanks{This work was supported in part by the National Natural Science Foundation of China (Grant 61871270), in part by the Shenzhen Natural Science Foundation under Grants JCYJ20200109110410133 and 20200812110350001, in part by the National Engineering Laboratory for Big Data System Computing Technology of China. \emph{(Corresponding Author: Dr. Xu Wang)}
			
			Jinlong Li, Xu Wang and Yu Zhou are with the College of Computer Science and Software Engineering, Shenzhen University, China, and also with Guangdong Laboratory of Artificial Intelligence and Digital Economy (SZ), Shenzhen University, Shenzhen, 518060, China. Email: (jinlong.szu@gmail.com; wangxu@szu.edu.cn; yu.zhou@szu.edu.cn).
			
			Zequn Jie, Xiaolin Wei and Lin Ma are with Meituan Inc., China. Email: (zequn.nus@gmail.com; weixiaolin02@meituan.com; forest.linma@gmail.com).
			
			{$^{\dagger}$These authors contributed equally to this work.}
			
			Copyright © 20xx IEEE. Personal use of this material is permitted. However, permission to use this material for any other purposes must be obtained from the IEEE by sending an email to pubs-permissions@ieee.org.
	}}

	\markboth{IEEE Transactions on Multimedia, xx 20xx}%
	{Li \MakeLowercase{\textit{et al.}}: WSSS via Progressive Patch Learning}
	%



	\maketitle
	
	
	\begin{abstract}
		Most of the existing semantic segmentation approaches with image-level class labels as supervision, highly rely on the initial class activation map (CAM) generated from the standard classification network. In this paper, a novel ``Progressive Patch Learning'' approach is proposed to improve the local details extraction of the classification, producing the CAM better covering the whole object rather than only the most discriminative regions as in CAMs obtained in conventional classification models. ``Patch Learning'' destructs the feature maps into patches and independently processes each local patch in parallel before the final aggregation. Such a mechanism enforces the network to find weak information from the scattered discriminative local parts, achieving enhanced local details sensitivity. ``Progressive Patch Learning'' further extends the feature destruction and patch learning to multi-level granularities in a progressive manner. Cooperating with a multi-stage optimization strategy, such a ``Progressive Patch Learning'' mechanism implicitly provides the model with the feature extraction ability across different locality-granularities. As an alternative to the implicit multi-granularity progressive fusion approach, we additionally propose an explicit method to simultaneously fuse features from different granularities in a single model, further enhancing the CAM quality on the full object coverage. Our proposed method achieves outstanding performance on the PASCAL VOC 2012 dataset (\textit{e.g.}, with 69.6$\%$ mIoU on the \textit{test} set), which surpasses most existing weakly supervised semantic segmentation methods. Code will be made publicly available here \url{https://github.com/TyroneLi/PPL_WSSS}.
	\end{abstract}
	
	
	%
	\IEEEpeerreviewmaketitle

	\section{Introduction}
	
	\IEEEPARstart{I}{mage} semantic segmentation is one of the fundamental tasks of computer vision which makes huge progress with the development of deep learning in recent years~\cite{chen2017deeplab,chen2014semantic,long2015fully}. However, training a segmentation network in a fully-supervised manner requires a large number of pixel-wise annotations which costs tremendous human-labor. To reduce the effort of human labeling, weakly-supervised semantic segmentation (WSSS) has been studied for years, and various types of weak annotations,~\textit{e.g.}, image-level ~\cite{wei2018revisiting,ahn2018learning,wei2017object,wang2018weakly,kolesnikov2016seed, huang2018weakly},  scribles~\cite{lin2016scribblesup,vernaza2017learning}, point-level~\cite{bearman2016s} and bounding box~\cite{dai2015boxsup,khoreva2017simple,papandreou1502weakly} labels, are considered in practice. Our work focuses on the most challenging one that only adopts image-level labels for WSSS. Compared to other types of weak labels, image-level labels can be obtained in a more convenient and free way, given the huge number of Internet images with image-level tags, and thus has great potential in practical use. 
	
	\begin{figure}
		\begin{center}
			\includegraphics[width=0.48\textwidth]{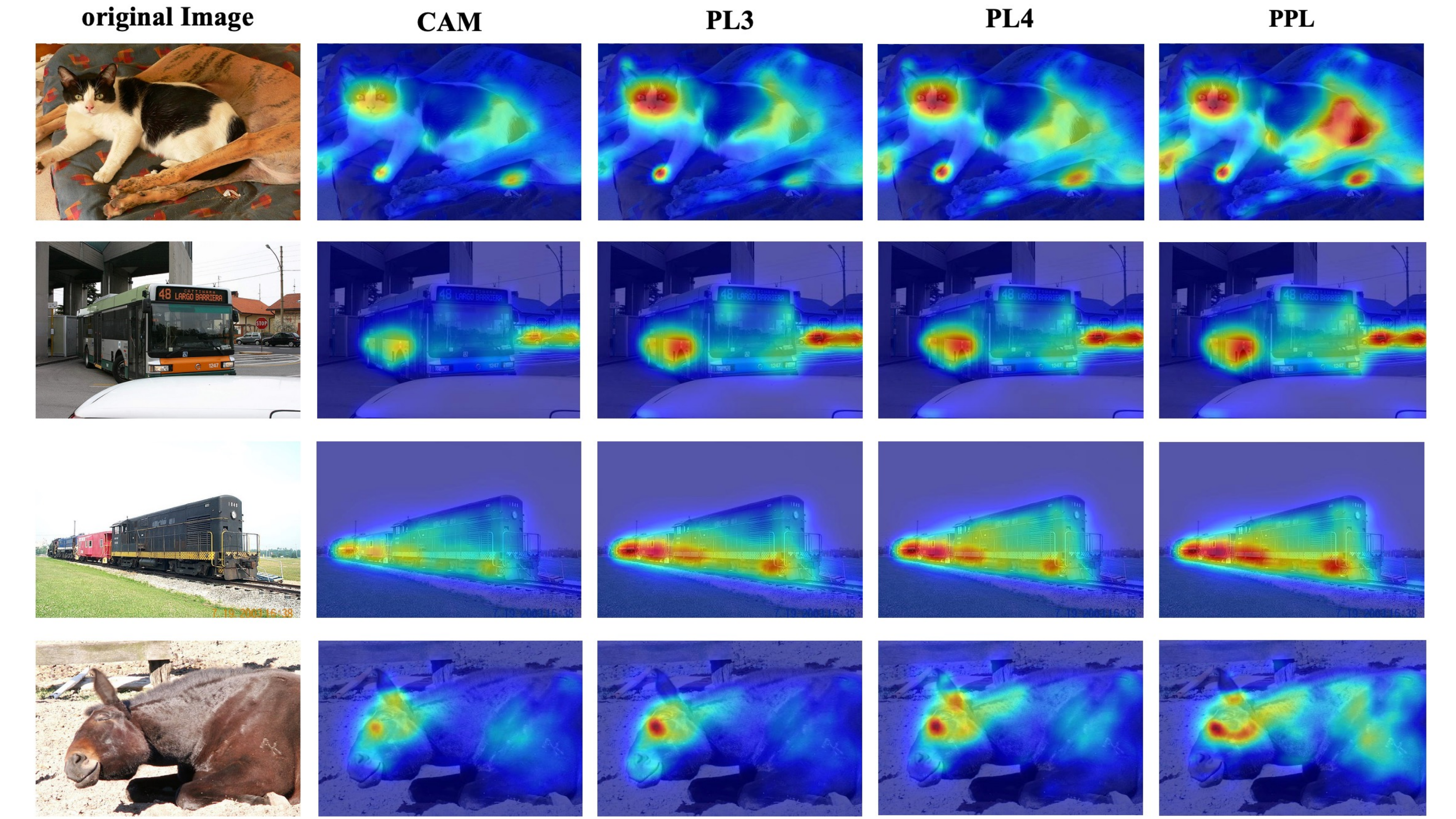}
		\end{center}
		\captionsetup{font={small}}
		\caption{{Visualization of the CAM map. CAM denotes the class activation map generated by the baseline model, PL3 and PL4  denotes the CAM map obtained by applying Patch Learning only in the beginning of ResNet Stage 3 and Stage 4, respectively.  PPL denotes  the CAM generated by implementing our full Progressive Patch Learning. Noting that our results are generated via proposed implicit method.}}
		\label{fig:fig_1}
		\vspace{-0.5cm}
	\end{figure}
	
	Most existing approaches follow the two-stage pipeline in WSSS, \emph{i.e.}, firstly obtaining initial class specific activation maps via image-level classification networks and refining the activation response masks to get pseudo annotation masks, and then training a segmentation network in a fully-supervised manner based on the generated pseudo labels. The Class Activation Map (CAM)~\cite{zhou2016learning} is widely utilized as the initial estimate of the pseudo ground truth masks. CAM is usually computed via a weighted sum over the feature maps in the last layer of the multi-label classification network. However,  there exists an inherent mismatch between the image-level classification task and the demand for dense pixel-level pseudo mask labels for training a fully-supervised segmentation network. Highlighted regions in CAM usually focus on the most discriminative parts of the  object and fail in capturing precise boundaries (see Fig.~\ref{fig:fig_1}), and thus deviating from the requirement of the semantic segmentation task. To recover more complete object areas from the initial response map, existing approaches rely on various techniques, \emph{e.g.}, dilated convolutions~\cite{wei2018revisiting},  hide-and-seek strategy~\cite{wei2017object, singh2017hide}, randomly dropping connections~\cite{lee2019ficklenet}, and online accumulation  strategy~\cite{jiang2019integral}. However, these approaches either perform complicated iterative training,  or still struggle in balancing  the recall of the foreground and the false-positives of the background. For example, \cite{wei2018revisiting} and \cite{lee2019ficklenet} require unsupervised saliency cues to filter false-positive regions.

	This paper tackles with the partial localization issue of CAM in a novel solution called \textbf{``Progressive Patch Learning'' (PPL)}. As image classification networks have strong high-level semantic representation learning ability, CAM at the last layer of a classification network typically attends only the most discriminative parts. To break up the global structure of an image and allow the model to explore the scattered discriminative local details, this paper proposes a novel ``Progressive Patch Learning'' solution based on feature destruction operation. Specifically,   feature maps at the deep layers which focus on global  features are deliberately destructed into several patches, and each patch  passes through the subsequent convolutional layers separately. Since high-level semantic features describing the whole object are broken down and absent to the classification network, the network has only access to the local features related to the low-level object parts. The lost of high-level global features makes it more difficult to make the correct classification given only a set of scattered object parts, and thus the model has to gather information from as many small patches as possible. Therefore more scattered small parts will contribute to the classification and get high activation response. In this way, the CAM would have more chances to cover the whole object.
	
	Going one step further, we find that simply breaking up global features in a fixed layer into patches  cannot produce optimal feature granularity for each  object instance, as in the third and  fourth columns of Fig.~\ref{fig:fig_1}. As known, objects in different scales and semantic levels may require different trade-offs between local details and global structures.  To achieve this goal, we propose a  multi-level feature destruction approach called Progressive Patch Learning (\textbf{PPL}), which breaks up the features at different granularities into patches progressively and trains the classification network in a multi-stage manner. More specifically, our PPL method starts from the more stable finer granularities which capture more low-/mid-level information, and gradually move onto the coarser ones. At each training step, feature destruction is shifted to a higher network block, and only the currently destructed layers are updated while all the previous layers are fixed. Fixing the non-destructed layers  effectively exploits locality-sensitive information from the last granularity learned at the previous training step, and allows the training to focus on the information at the current granularity. With the progressive multi-level patch learning mechanism, features at diverse granularities are implicitly fused progressively, providing model sensitivity on different locality-granularities and better coverage of different objects on CAM (see Fig.~\ref{fig:fig_1} fifth column).
	
	Furthermore, we propose a more explicit multi-level granularity fusion approach based on the progressive patch learning. In such an alternative solution, the classification network with different feature destruction layer positions are shared and simultaneously trained, and their respective features are finally concatenated to generate the fused feature. Benefiting from the explicit feature fusion, better coverage of object instances on CAM is naturally achieved. Both the implicit and explicit versions of our approach are allowed to train the classification model end-to-end without using any extra information,~\textit{e.g.}, pre-computed saliency map prior knowledge, or pre-computed superpixels such as~\cite{fan2020learning}, making it quite convenient to use in practice.

	We conduct extensive experiments on PASCAL VOC 2012 dataset~\cite{everingham2010pascal} to demonstrate the effectiveness of our approach, aiming to generate better initial CAMs to locate objects. As a result,  our approach leads to an outstanding performance for the final WSSS results. Moreover, we perform extensive ablation studies and experimental analyses to show the intermediate benefits brought by our method, shedding light on the future works on this task.
	
	Overall, the main contributions of this paper are summarized as follows.
	\begin{itemize}
		\item We propose a novel patch learning method that processes each image feature patch in parallel with convolution blocks and then gathers them to produce the enhanced feature with a stronger sensitivity to local details, which alleviates the partial localization issue of the CAM approach. 
		\item  We propose a novel  training strategy where patch learning is performed from bottom to top layers progressively,  implicitly enabling the model to learn local details at different granularities simultaneously.
		\item We further propose an explicit multi-granularity fusion approach that trains a classification network which applies feature destruction at different positions of the network simultaneously, without multi-stage granularities fusion steps.
		\item We present extensive experiments and analysis to validate the effectiveness of our approach. The proposed approach improves the quality of initial CAMs and obtains competitive WSSS performance.
	\end{itemize}
	
	
	
	\section{Related Work}
	In this section, we discuss approaches for WSSS adopting image-level labels. Due the tremendous need for pixel level image mask labeling, WSSS have been widely studied for semantic segmentation task. The WSSS task with image-level weak labels needs the least human annotation labor, but it only indicates the presence or the absence of objects in an image, resulting in an inferior segmentation performance when compared to the fully supervised counterparts. We describe the methods that focus on the initial CAM generation, refinement for the completeness of the pseudo ground truth masks and some related works that share similar operation with ours.
	
	\subsection{Initial CAM Generation for WSSS} 
	
	Obtaining more complete target object discriminative regions is the most important technique in the subsequent refinement training stage or semantic segmentation model training with only image-level supervisions. Since providing reliable priors to specific real object localization in image, the technique of CAM~\cite{zhou2016learning} is widely utilized among the WSSS community to locate the discriminative object regions for classification prediction. CAM discovers the related discriminative regions with respect to each individual target class. CAM utilizes the contribution of each hidden unit in a neural network to the final classification score, allowing the hidden units which make large contributions to be identified. Meanwhile, it can only cover some parts of the target object leading to sparse and incomplete initial CAM seeds. Grad-CAM~\cite{selvaraju2017grad} is a generalization of the CAM that uses generalized weights to derive the score map, cooperating with dense gradients of the hidden units. {Score-CAM~\cite{wang2020score} computes the weight of each activation map through its forward passing score instead of the gradient  and its acceleration method Group-CAM~\cite{zhang2021group} uses data-driven methods to {generate CAM interpretation}. Layer-CAM~\cite{jiang2021layercam} uses diverse feature maps coming from different CNN layers to generate CAM maps.} Differently, Zhang \textit{et al.}~\cite{zhang2018top} proposes Excitation Backprop method to back-propagate in the network through a hierarchical structure to find out the discriminative regions. Zhang \textit{et al.}~\cite{zhang2013representative}  proposes an algorithm that learns the distribution of spatially structural superpixel sets from image-level labels. And Zhang \textit{et al.}~\cite{zhang2019decoupled} proposes a decoupled spatial neural attention network to generate pseudo-annotations by localizing the discriminative parts of the object region. Several approaches have been devoted to alleviating the incomplete seed problem via erasure strategy. AE-PSL~\cite{wei2017object} and SeeNet~\cite{hou2018self} erase the most discriminative target object regions and forces the network to identify other non-discriminative regions, to make CAM cover more different regions. Hide-and-Seek~\cite{singh2017hide} hides the discriminative region of an object from the input image level, forcing the model to seek more non-discriminative parts. Some variants~\cite{li2018tell,zhang2018adversarial} further propose an adversarial erasing method in an end-to-end training manner to progressively discover and expand target object discriminative regions. However, the expansion could also introduce the background regions, leading to inaccurate attention maps. Therefore, to refine the generated pseudo mask labels, saliency cues have been widely exploited to estimate the background regions~\cite{lee2019ficklenet}, ~\cite{wang2020self}, ~\cite{chang2020weakly}, ~\cite{fan2020cian}. 
	
	Recently, Ficklenet~\cite{lee2019ficklenet} performs stochastic feature selection at both training and inference phases, and aggregates them to obtain the initial CAM seed. SEAM~\cite{wang2020self} utilizes the equivariance of the ideal segmentation function for network learning, forces the network to learn consistent CAMs among different transformed input images. SC-CAM~\cite{chang2020weakly} adopts sub-category exploration to enhance feature representations and improve the initial response map via iterative clustering method. They all utilize the self-supervised learning mechanism. Zhou \textit{et al.}~\cite{zhou2020sal} proposes two losses, selection loss and attention loss, to alleviate the problem that the generated pseudo image mask labels inherently involve much noisy. Sun \textit{et al.}~\cite{sun2020mining} and Fan \textit{et al.}~\cite{fan2020cian} capture information shared between several images by considering cross-image semantic similarities and differences. MDC~\cite{wei2018revisiting} utilizes multiple convolutional layers with different dilation rates to expand the original activated regions in CAM. The recently proposed OAA~\cite{jiang2019integral} accumulates attention maps at different training epochs and introduces integral attention learning to acquire enhanced attention maps. However, OAA~\cite{jiang2019integral} may introduce undesired attention regions due to training instability in the early stage. {LVW~\cite{ru2021learning} proposes to enforce the model to learn local visual word labels, thus enabling CAM to cover more object extents. On the other hand, LVW proposes HSPP to aggregate multi-scale features and combine GAP and GMP for better object extent localization.} More recently, some works study an end-to-end network for the joint training of classification and segmentation. For example, S-Stage~\cite{araslanov2020single} and RRM~\cite{zhang2020reliability} explore an end-to-end training pipeline to save training time with the cost of lower segmentation performance.
	
	Unlike complicated erasing strategies during network training or iterative update after inferring the initial CAM seed at the training step, our approach aims to equip the model with the ability to cover complete object regions via exploring local information by patch-level mining. {Our approach also differs from DCSM~\cite{shimoda2016distinct} and MDC~\cite{wei2018revisiting} not only in the way of controlling the receptive field, but also in  how to aggregate the activation maps from different receptive fields into the final single map serving as the pseudo segmentation mask. Most of the mentioned works (e.g., DCSM~\cite{shimoda2016distinct} and MDC~\cite{wei2018revisiting}) achieves this by explicit hand-crafted summation. In contrast, our implicit fusion version directly outputs a single CAM after the progressive training where features at diverse granularities are implicitly fused gradually. As such, the model is equipped with sensitivity on the details of different locality-granularities and the final single CAM has a better coverage of different objects.}
	
	\subsection{Feature Response Refinement for WSSS}
	
	Another type of approaches aim to enlarge the very initial CAM seeds to cover more regions. Numerous approaches~\cite{kolesnikov2016seed,wang2018weakly,huang2018weakly,ahn2018learning,ahn2019weakly,chen2020weakly,fan2020learning}  refine the original parts by expanding the incomplete object regions in the activation map. SEC~\cite{kolesnikov2016seed}  uses three principles, \emph{i.e.}, seed, expand, and constrain, to refine CAMs, which is followed by many other works. MCOF~\cite{wang2018weakly} utilizes a bottom-up and top-down framework which alternatively expands object regions and optimizes the segmentation model. DSRG~\cite{huang2018weakly} dynamically updates the initial localization seed masks via applying a seeded region growing strategy during the training of the segmentation network. Other approaches are proposed via affinity learning. For example, AffinityNet~\cite{ahn2018learning} trains another network to learn the similarity between pixels, and obtains a transition matrix and multiplies with CAM map several times to expand respective response maps to propagate local response to the neighbors. IRNet~\cite{ahn2019weakly} generates a transition matrix from the boundary of class activation map and finally applies this method to weakly-supervised instance segmentation. ICD~\cite{fan2020learning} learns pixel affinity only to enhance feature representations for object estimation. BE~\cite{chen2020weakly} also explores object boundaries via an affinity learning to provide object boundary constraint to refine the semantic segmentation output. Meanwhile, external saliency models~\cite{jiang2013salient,hou2017deeply} are adopted to estimate background clues which implicitly introduces additional pixel-level annotation required for training such saliency models. Anti-Adv~\cite{lee2021anti} utilizes an anti-adversarial manipulation method to expand the most discriminative regions in the initial CAMs to other non-discriminative regions. EDAM~\cite{wu2021embedded} uses self-attention layer to identify class specific attention maps via collaborative multi-attention training.
	
	\begin{figure*}[t]
		\begin{center}
			\includegraphics[width=0.98\textwidth]{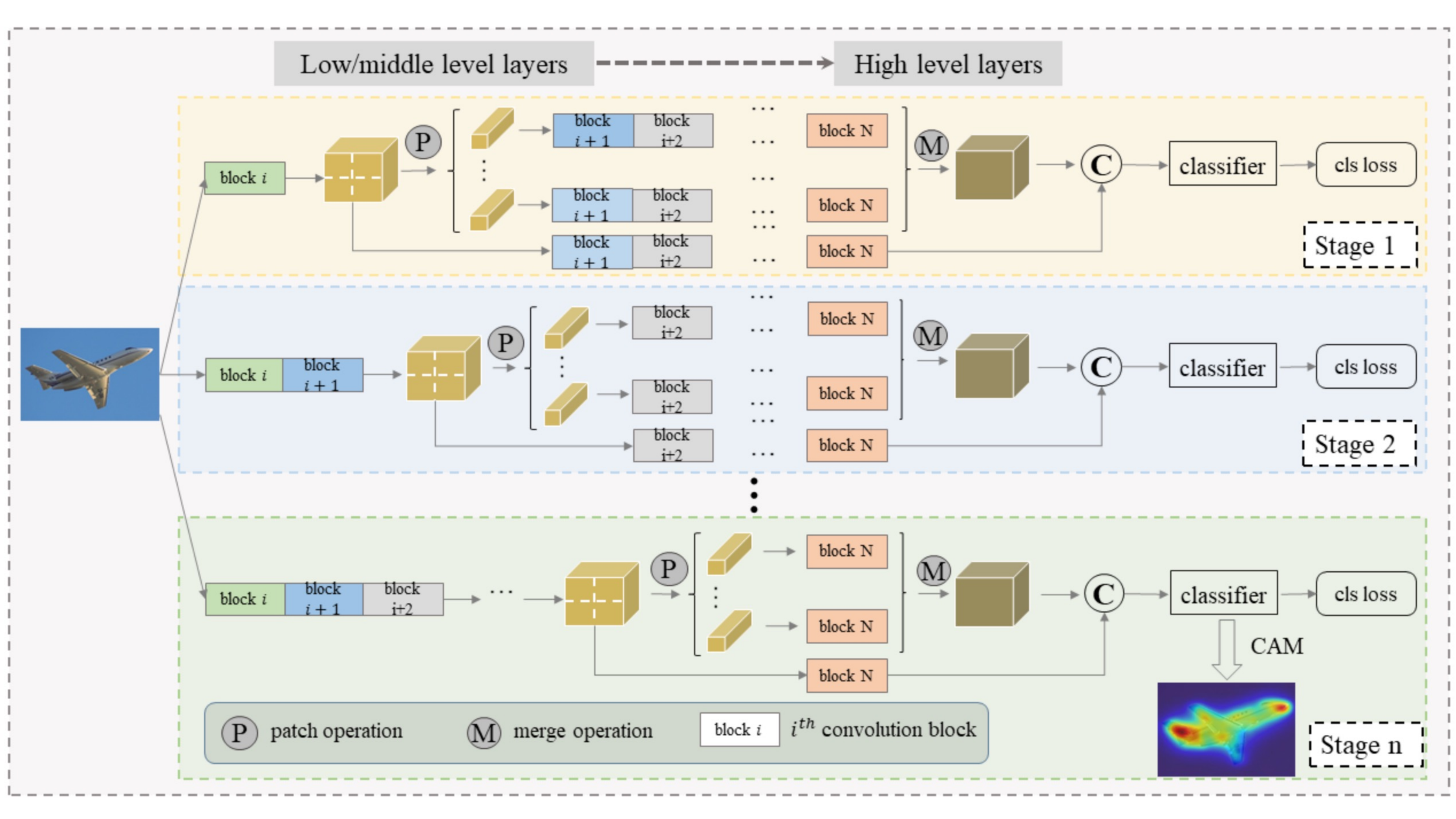}
		\end{center}
		\captionsetup{font={small}}
		\caption{Our proposed Progressive Patch Learning pipeline. \textbf{P}, \textbf{M} and \textbf{C} denote the proposed Patch Operation, Merge Operation and Feature Concatenation, respectively. Different color denotes that patch learning is applied at different positions of the network. Notice that our approach only adopts one single classification model which is progressively trained.
		}
		\label{fig:fig_2}
		\vspace{-0.2cm}
	\end{figure*}
	
	\subsection{Patch-based Learning}
	
	Recently, patching image into several partial tiles is widely used in self-supervised learning~\cite{noroozi2016unsupervised,doersch2015unsupervised,noroozi2018boosting,carlucci2019domain,xie2021detco}. These methods destruct image into patches and apply contrastive learning to enforce network to cluster the same instances and push away the distinct instances. Meanwhile, such a training mechanism forces the network to focus more on local fine-grained features, enabling the network to learn more sophisticated local features and relationships among patches.  Different from these methods, our method utilizes destruction learning in the  feature level to deliberately damage the large receptive field to make the network mine more local non-discriminative features during optimizing the final classification task, and thus locate more complete target object regions.

	
	\section{Proposed Method}
	
	In this section, we present our proposed PPL approach which aims to improve the quality of the initial CAM seed of multi-label classification model, \textit{i.e.}, to cover more complete object regions. As shown in Fig.~\ref{fig:fig_2}, our framework trains a classification network in a multi-stage manner. The model is trained from the finer granularities which capture more low-/mid-level information, and gradually moved onto coarser ones. At each training step, feature destruction is shifted to a higher network block, and only the currently destructed layers are updated while all  the  previous layers are fixed. Such a training strategy exploits locality-sensitive information from the last granularity learned at the previous training step, and allows the training to focus on the information of the current granularity. Finally, the CAM obtained in the last stage of training which attends discriminative regions across different granularities is utilized as the initial pseudo label.

	We firstly describe our preliminary observation and the motivation of our approach in Sec.~\ref{sec2_1}, showing that our method could alleviate the issue of partial object localization for the naive CAM method. Secondly, we illustrate our proposed Patch Learning in Sec.~\ref{sec2_2} (including Patch Operation and Merge Operation). Thirdly, we present the implicit and explicit fusion strategies in Sec.~\ref{sec2_3} and Sec.~\ref{sec2_4}, respectively. Finally, we explain our initial CAM seed generation method in Sec.~\ref{sec2_5} and the final semantic segmentation model training scheme in Sec.~\ref{sec2_6}.
	
	\subsection{Observation}
	\label{sec2_1}
	
	\begin{figure}
		\centering
		\includegraphics[width=0.48\textwidth]{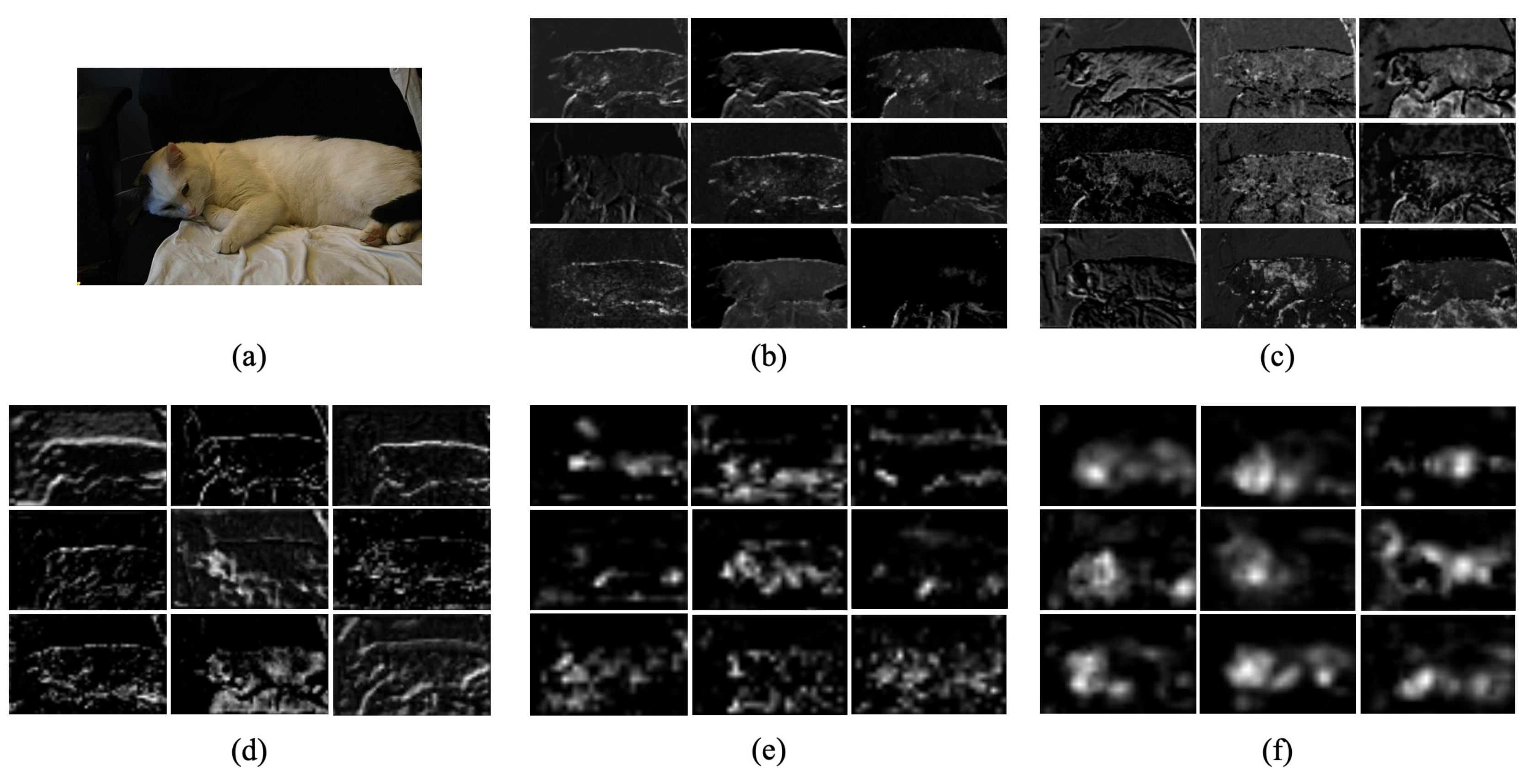}
		\captionsetup{font={small}}
		\caption{Example of feature visualization. We visualize the feature maps randomly selected from channels from each Residual Block Stage of the common ResNet model. (a) denotes the original image, (b) (c) (d) (e) (f) are visualization demos after ResNet Stem, Stage 1, Stage 2, Stage 3, Stage 4 block, respectively. Best view zoom in.}
		\label{fig:fig_3}
	\end{figure}
	
	WSSS approaches usually train a multi-label classification network, and then use CAM methods to extract the last layer features to obtain the coarse segmentation label. However, due to the hierarchical structure of CNN model, it gradually focuses on the most discriminative object regions from bottom to top layers. The resulting low-quality CAM map usually hampers the following steps such as refinement or segmentation model training. We visualize some feature responses from different layers from the common classification (ResNet) model at Fig.~\ref{fig:fig_3} to demonstrate this phenomenon. As can be observed that when the network layer goes from low/mid-level to high-level layer blocks, the feature responses change from the stable low-level details to the  high-level discriminative parts. This results from the evolving of both the semantic level and receptive field of network layers. Since top layers have larger receptive fields and higher levels of semantic information, they tend to choose the most informative regions for the classification among the large receptive input regions. Usually the most discriminative object regions captured by the network are sufficient to make the final correct category prediction, and the less discriminative ones are ignored. However, such partial object activation regions are not good enough for the   dense pixel-wise semantic segmentation task without correct pixel-level supervisions. We thus propose a ``Progressive Patching Learning'' solution to address this issue. 
	
	Therefore, the key to address the partial localization issue of CAM is to constrain the receptive fields of top layers and thus break up the global structure of the learned object features in these layers. In this case, the network has to learn to aggregate finer local object cues to satisfy the classification task. Therefore, more complete object regions could be activated and the partial activation issue could be alleviated to provide better pseudo segmentation mask supervisions for the final semantic segmentation model training.

	\subsection{Patch Learning}
	\label{sec2_2}
	
\begin{figure}
		\centering
		\includegraphics[width=0.48\textwidth]{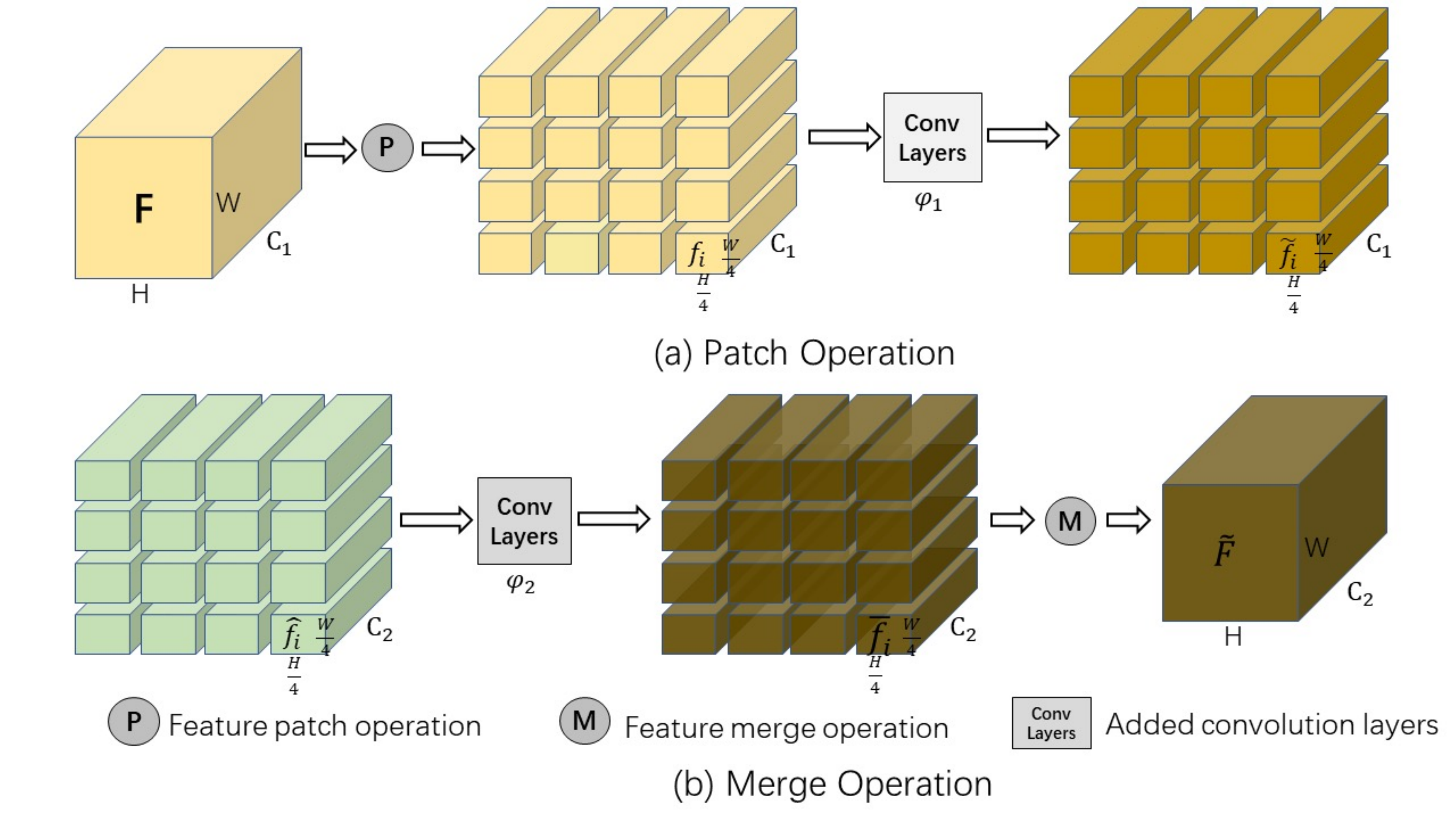}
		\caption{Example of the proposed Patch Operation and Merge Operation with $4 \times 4$ patches along the spatial dimension. $H$, $W$ and $C_1$ denote the height, width and channels of the input feature maps respectively. $\varphi_1$ is the shared $C_1 \times k_1 \times k_1$ convolution layer transformation after the  destruction of the feature maps. $\varphi_2$ is the shared $C_2 \times k_2 \times k_2$ convolution layer transformation before  merging the patch-wise feature maps.}
		\label{fig:fig_4}
	\end{figure}

	Our proposed Patch Learning aims to destruct the global structure features highlighting only the discriminative object regions and forces the network to seek as more local small patches as possible. To this end, we design two operation modules as shown in Fig.~\ref{fig:fig_2} marked as P (Patch Operation) and M (Merge Operation). Specifically, given the input feature maps $\mathbf{F} \in \mathbb{R}^{C_{1} \times H \times W}$ obtained from a particular convolution block, where $H \times W$ is the spatial 
	resolution and $C_{1}$ is the number of channels,
	the Patch Operation firstly slices the input feature maps $\mathbf{F}$ into $K \times K$ patches along the spatial dimension. As shown in Fig.~\ref{fig:fig_4}, for the $j$-th patch denoted as $\mathbf{f}_j \in \mathbb{R}^{C_{1} \times \lceil \frac{H}{K} \rceil \times \lceil \frac{W}{K} \rceil}$, a shared convolution layer $\varphi_1$ with kernel size $C_{1} \times 3 \times 3$ is utilized to re-encode feature patches since the global feature structure is destroyed. We thus obtain a new feature for each of the $j$-th patch, \emph{i.e.}, $\tilde{\mathbf{f}}_{j}=\varphi_1(\mathbf{f}_j)$. Then all the patch features are processed through all the following layers independently. After that, the Merge Operation is introduced to collect the destructed patches to form a new feature $\tilde{\mathbf{F}}$. Specifically, each feature patch is processed by a shared convolution layer $\varphi_2$ with kernel size $C_{2} \times 3 \times 3$, and then gathered along the spatial dimension  and spliced together to generate a new feature map $\tilde{\mathbf{F}}$ according to their sliced order in Patch Operation. Meanwhile, the non-destructed feature map $\mathbf{F}$ is also processed by all the convolution blocks with the  weights shared with  those  applied to the patches, to obtain $\mathbf{\hat{F}}$ with the same dimension as $\tilde{\mathbf{F}}$. This guarantees that $\mathbf{\hat{F}}$ and $\tilde{\mathbf{F}}$ are at the same semantic level and can be directly fused later. $\mathbf{\hat{F}}$ and $\tilde{\mathbf{F}}$ are complimentary where the global and local information are more focused respectively, and thus feature fusion of them will be beneficial to attend objects of diverse scales and semantic levels. Therefore, we concatenate the feature maps $\mathbf{\hat{F}}$ and $\tilde{\mathbf{F}}$ as the new feature $[\mathbf{\hat{F}}, \tilde{\mathbf{F}}] \in \mathbb{R}^{2C_{2} \times H \times W}$ along channel dimension to feed into the final  classification layer.
	
\begin{figure*}
		\centering
		\includegraphics[width=0.98\textwidth]{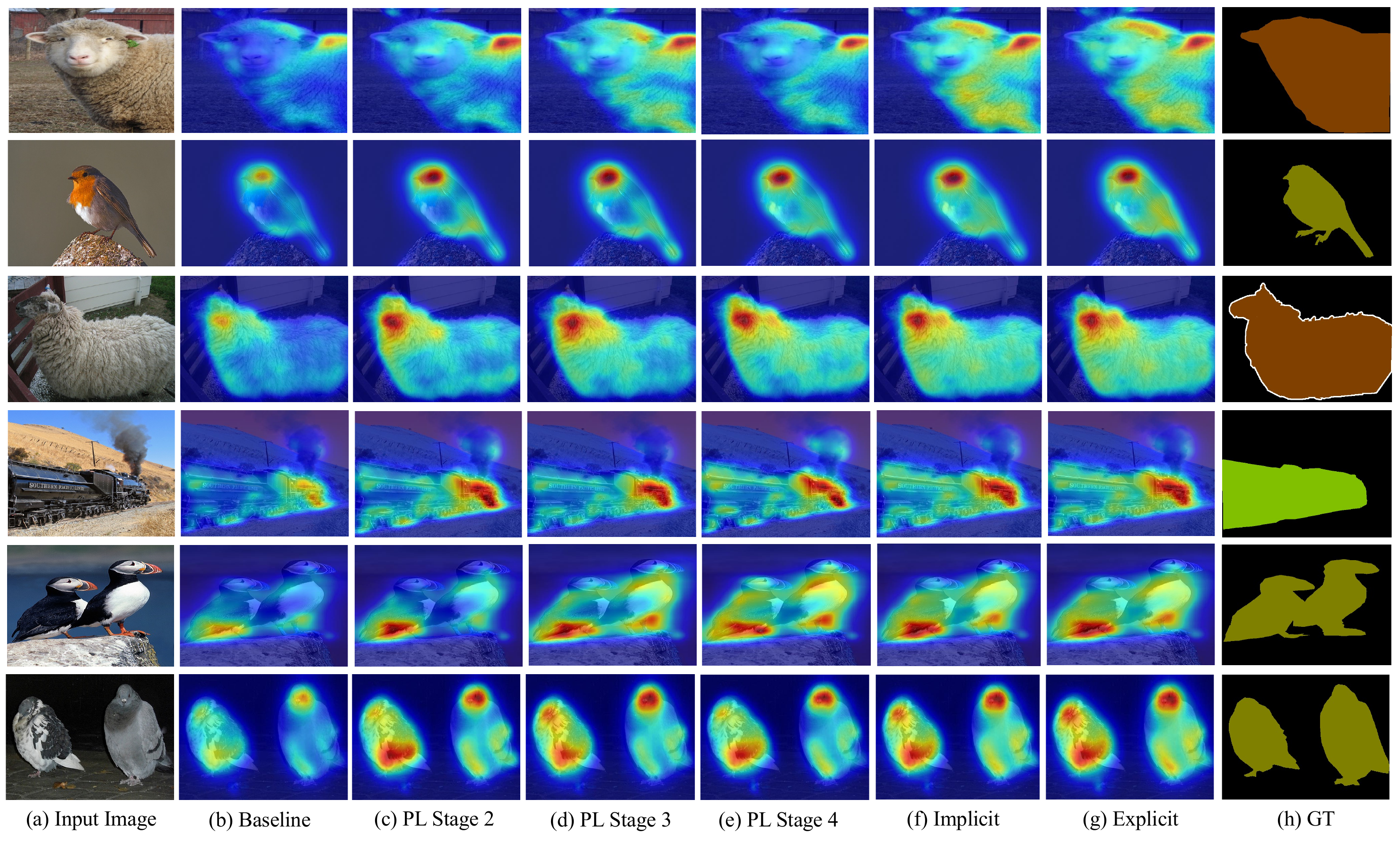}
		\captionsetup{font={small}}
		\caption{Visualizations of the CAM map comparison. ``PL'' means the proposed Patch Learning. (c) denotes applying Patch Learning before ResNet-50 Stage 2  only, while (d)(e) denotes the same implementation as (c) except the different destruction position. (f) illustrates the full Progressive Patch Learning  (implicit) result, while (g) is the explicit result.}
		\label{fig:vis_cam}
	\end{figure*}

	Such a patch learning mechanism emphasizes more local features as each distributed patch is processed by the network independently before final aggregation. Since the global information is damaged, the model would be forced to enhance its sensitivity to the local parts. In this way, the non-discriminative regions could also stand out and be highlighted in a patch as well. Naturally, the CAMs can cover more positive foreground local parts, and thus serves as better initial ground truth masks for the semantic segmentation training. Fig.~\ref{fig:vis_cam} shows the improvement on CAM by our proposed patch learning method compared with the original CAM.
	
	\begin{figure*}[t]
		\begin{center}
			\includegraphics[width=0.98\textwidth]{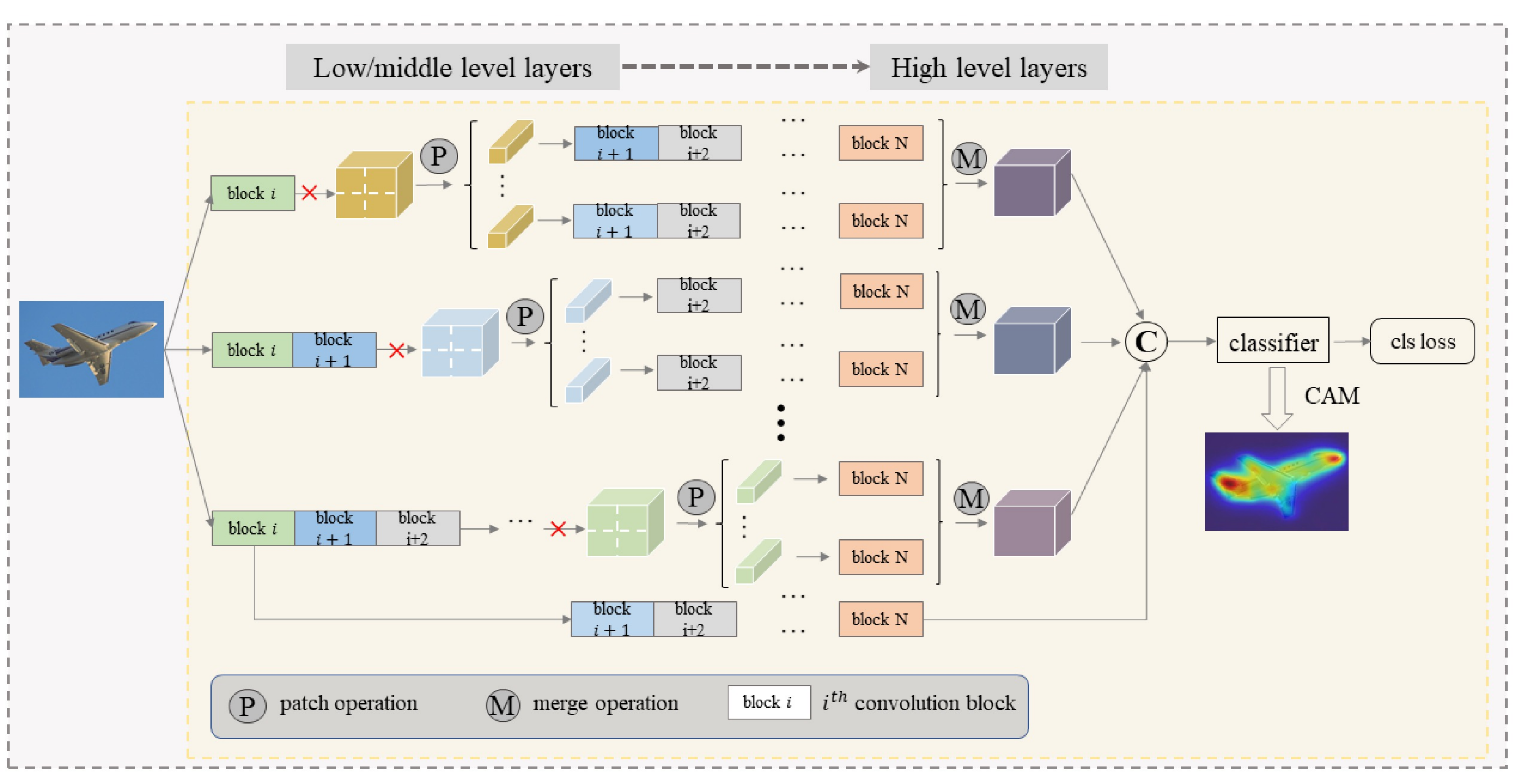}
		\end{center}
		\captionsetup{font={small}}
		\caption{Our proposed Explicit Patch Learning pipeline. \textcolor{red}{\tiny \XSolid} denotes detaching gradient back-propagation towards lower layers. \textbf{P}, \textbf{M} and \textbf{C} denote proposed Patch Operation, Merge Operation and Feature Concatenation, respectively. The two-dimensional rectangle box denotes the same share convolution blocks.}
		\label{fig:fig_explicit}
		\vspace{-0.2cm}
	\end{figure*}
	
	\begin{algorithm}[h]
		\caption{Implicit Progressive Patch Training Pipeline}  
		\label{alg:algorithm1}  
		\begin{algorithmic}[1]  
			\Require  
			$i$: the  convolution block index where the first Patching Operation is performed;
			$N$: total convolution blocks number.
			\Ensure  
			optimal High Quality CAM Result.
			\State initialize the network with ImageNet pretrained weights;
			\State apply Patch Learning method at the $i$-th conv block; 
			\State train the classification model, obtain the trained model denoted as $M$;
			\If {$i == N$}
			\State goto $15$-th step;
			\EndIf
			\Repeat  
			\State $i$++;  
			\State initialize network backbone weights from $M$; 
			\State fix all the layers before $i$;
			\State embed Patch Learning method at $i$-th conv block; 
			\State train the classification model;
			\State replace $M$ with the newly trained model in the $12$-th step;  
			\Until{($i == N$)}  
			\State infer CAM results utilizing the final classification model.
		\end{algorithmic}  
	\end{algorithm}  
	
	
	\subsection{Implicit Fusion by Progressive Training}
	\label{sec2_3}
	
	To further mine features at diverse locality granularities, we  propose a progressive learning strategy. Our progressive training starts from the bottom layer with finer granularities, and gradually move onto top layer with coarser granularities. At each training step, feature destruction is shifted to a higher network block, and only the currently destructed layers are updated while all the previous layers are fixed. We illustrate this progressive training strategy in Algorithm~\ref{alg:algorithm1}.
	
	Freezing the non-destructed convolution blocks enforces the network to fully exploit the locality-sensitive cues from the last granularity learned at the previous training step, and focus only on the object parts at the current larger granularity. Training the network with such a strategy gradually accumulates different local feature granularities from bottom to top layers which could alleviate partial localization issue. We adopt such a training scheme to fuse  feature responses from different granularities inside the hierarchical CNN structure implicitly.
	
	\subsection{Explicit Fusion}
	\label{sec2_4}
	Moreover, we also explore a more explicit and stronger way of multi-granularity fusion to further improve the CAM quality. Specially, we directly embed ``Patch Learning'' into different convolution blocks and concatenate the corresponding features to obtain the final fused feature, and update the convolution blocks applied to the destructed patches from all the granularities simultaneously during training. Specifically, we firstly train a non-destructed classification network on image-level annotations. After that, we introduce a multi-branch shared network, where each branch has different destruction layer positions (see Fig.~\ref{fig:fig_explicit} ) with the weights initialized by the non-destructed classification network. And the corresponding updated gradients are only allowed to flow backwards the convolution blocks where the corresponding Patch Operation is applied. In the Merge Operation, the collected features $\tilde{\mathbf{F}}_{l}$ (where $l=1,2,...,n$) from $n$ granularities are firstly obtained, and then we directly concatenate all the $\tilde{\mathbf{F}}_{l}$ with $\mathbf{\hat{F}}$ to form the final feature $\mathbf{F_{final}}=[\mathbf{\hat{F}}, \tilde{\mathbf{F}}_{1},\tilde{\mathbf{F}}_{2},...,\tilde{\mathbf{F}}_{n}] \in \mathbb{R}^{(C_{2}+C_{2,1}+C_{2,2}+\dots+C_{2, n}) \times H \times W}$. Here $C_{2}$ is the channel number of $\mathbf{\hat{F}}$, and $C_{2,l}$ is the channel number of $\tilde{\mathbf{F}}_{l}$.  Finally, $\mathbf{F_{final}}$ is fed into a single multi-label classification layer. The whole classification network is trained with only image-level labels by using a multi-label soft-margin classification loss $\mathbf{\mathcal{L}}$ between the ground truth label $y$ and prediction probability distribution $\hat{y}$ for the loss computation as:
	
	\begin{equation}
			\mathcal{L}(\hat{y}, y) = -\frac{1}{C} * \sum_{i}{y[i]*\log((1 + e^{-\hat{y}[i]})^{-1}) + \\ 
				(1-y[i])*\log({\frac{e^{-\hat{y}[i]}}{1+e^{-\hat{y}[i]}}})},
	\end{equation}
	where $C$ is the total number of training classes, $i$ represents the $i^{th}$ training sample in the minibatch, $y[i]$ is the ground truth label of $i^{th}$ class with the value of either 0 or 1.
	
	The explicit fusion model needs to be trained only once, which is more  efficient than the multiple training steps of the implicit fusion one, with the cost of higher memory consumption of the whole model.  
	
	\subsection{CAM Generation for Label Synthesis}
	\label{sec2_5}

	CAM~\cite{zhou2016learning} identifies the highlighted  object regions contributing the most to the final image-level classification, and thus can be used as the rough localization result of the target object. The obtained CAM is usually used as the pseudo ground truth mask the train the later semantic segmentation model.
	
	For our proposed implicit PPL method, we obtain two CAM seeds $M_{\mathbf{\hat{F}}}^c$ for $\mathbf{\hat{F}}$ and $M_{\tilde{\mathbf{F}}}^c$ for $\tilde{\mathbf{F}}$ from the network. Specifically, we compute them via:
	\begin{equation}
		M_{\mathbf{\hat{F}}}^c (x, y) = \sum_{k} \omega_{\mathbf{\hat{F}}}(k, c) \cdot \mathbf{\hat{F}}(k, x, y),
		\label{eq1}
	\end{equation}
	where $\omega_{\mathbf{\hat{F}}}(k, c)$ is the classifier weight for the category $c$ and channel $k$, $(x, y)$ is the spatial location of feature $\mathbf{\hat{F}}$. Then the activation map is further normalized by the maximum value in $M_{\mathbf{\hat{F}}}^c$ and thresholded by a value $\tau$. We obtain the CAM seed $M_{\mathbf{\tilde{F}}}^c$ of the class $c$ contributed by the feature maps $\tilde{\mathbf{F}}$ similar to Eq.~(\ref{eq1}). The initial CAM $M_{c}$ is defined as the average of $M_{\mathbf{\hat{F}}}^c$ and $M_{\mathbf{\tilde{F}}}^c$.
	
	For the proposed explicit Progressive Patch Learning approach, we compute CAM seed for each concatenated feature branch, similar to the implicit one, and obtain each respective CAM $M_{\mathbf{F}_l}^c$. We then obtain the initial CAM result $M_{c}$ which is defined as the average of $M_{\mathbf{F}_l}^c$ and $M_{\mathbf{\hat{F}}}^c$ (where $l$ = 1, 2, ..., n).
	After that, the same threshold value as the implicit one is adopted to process $M_{c}$. 
	
	\subsection{Segmentation Networks Training}
	\label{sec2_6}
	
	Since CAMs are obtained from the down-sampled features computed via the classifier, it could only localize the target object regions coarsely and  hardly satisfies the requirement of high-quality mask supervision for the semantic segmentation model training. Hence,  many methods  try to improve the initial CAM seeds with diverse subsequent refinement methods~\cite{ahn2019weakly,ahn2018learning,chang2020weakly,liu2020weakly,zhang2020causal}. SC-CAM~\cite{chang2020weakly} adopts AffinityNet~\cite{ahn2018learning}; MBMNet~\cite{liu2020weakly} and CONTA~\cite{zhang2020causal} use IRNet~\cite{ahn2019weakly}. Here we also apply the seed refinement method to the initial coarse CAM seed $M_{c}$ to further obtain refined  pseudo segmentation labels. Then we use the resulting  pseudo segmentation labels as the supervision to train a semantic segmentation network.
	
	\section{Experimental Results}
	In this section, we firstly describe the experiments setups in Sec.~\ref{sec3_1}, implementation details in Sec.~\ref{sec3_2}, and present our evaluation results on CAM map quality and compare it with previous state-of-the-art weakly supervised segmentation methods in Sec.~\ref{sec3_3}. Finally, we conduct a series of experiments to demonstrate the effectiveness of our proposed approach, including our analysis of pseudo labels and ablation study in Sec.~\ref{sec3_4}.
	
	\begin{figure}
		\centering
		\includegraphics[width=0.49\textwidth]{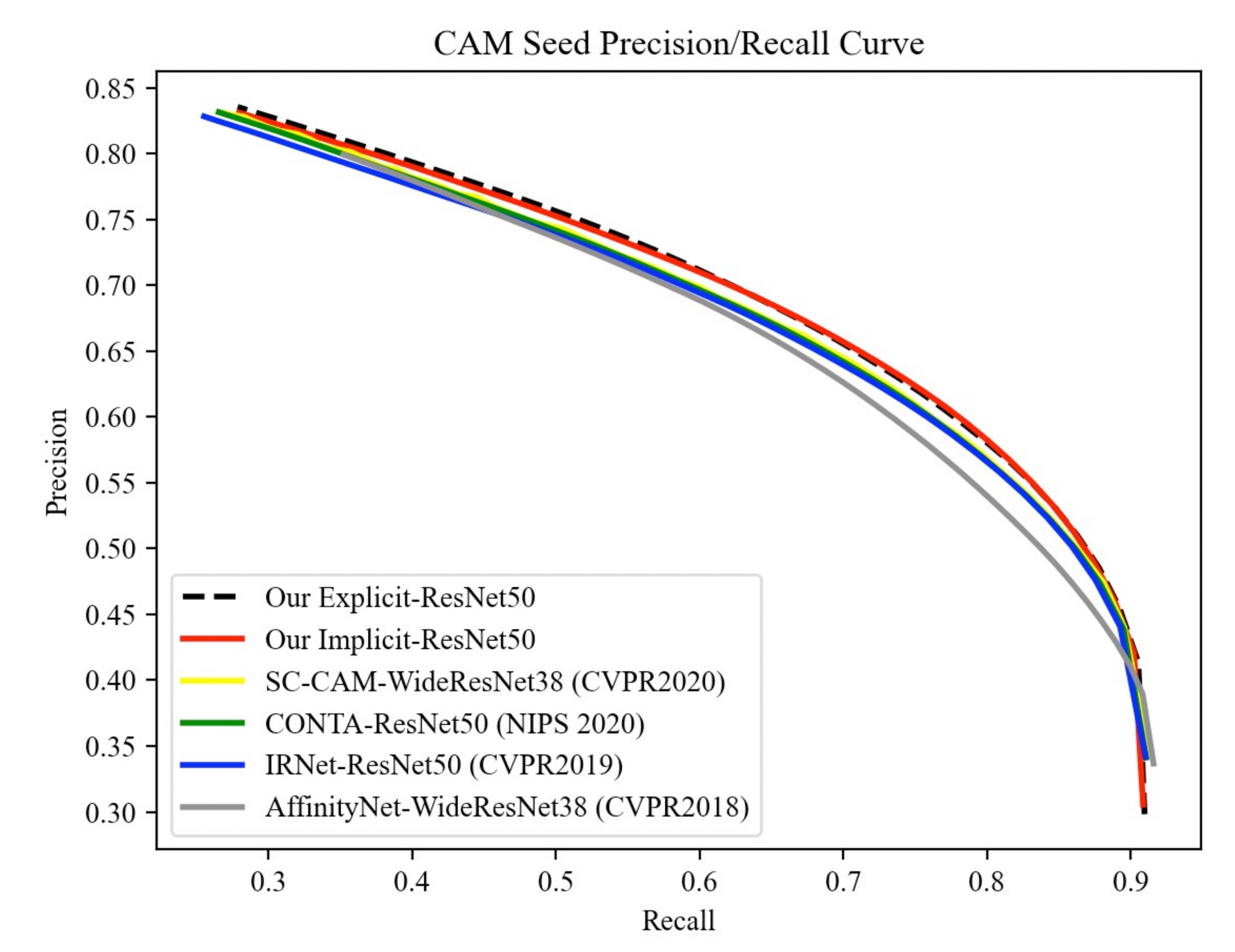}
		\captionsetup{font={small}}
		\caption{Precision and Recall Curve comparison with other SOTAs. This is obtained via setting various threshold values to calculate the corresponding precision and recall. Here we only make comparison with the methods applying refinement procedure for the fairness.}
		\label{fig:fig_pr}
	\end{figure}
	
	\begin{figure*}
		\centering
		\includegraphics[width=1.0\textwidth]{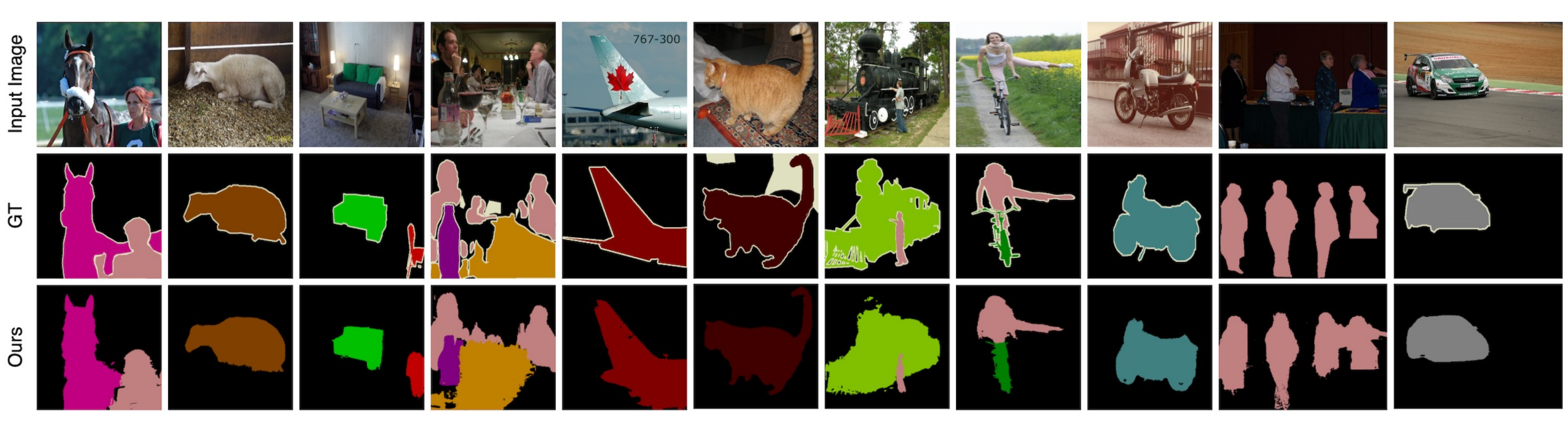}
		\centering
		\captionsetup{font={small}}
		\caption{Visualisation of segmentation results. From top to bottom are the raw image, ground truth and our segmentation results on PASCAL VOC 2012 \textit{val} set.}
		\label{fig:vis_seg}
	\end{figure*}
	
	\subsection{Dataset and Evaluation Metrics}
	\label{sec3_1}
	
	This section demonstrates the experiment setups we use and the effectiveness of our approach with comparisons to existing state-of-the-art WSSS methods on the PASCAL VOC 2012 segmentation benchmark~\cite{everingham2010pascal} with 21 class categories, \textit{i.e.}, 20 foreground objects and 1 background. Following the common protocol as previous WSSS methods, we adopt additional annotations from SBD~\cite{hariharan2011semantic} to construct an augmented \textit{train} set with 10582 images based on 1464 training images. Our proposed model is trained only with the image-level classification labels. We evaluate the quality of initial CAM seeds on the \textit{train} set without augmentation. To compare our results with other methods on semantic segmentation task, we adopt 1449 images in the \textit{val} set and 1456 images in the \textit{test} set for evaluation. No additional data is being adopted in our entire train/test pipeline. The MS COCO 2014 dataset~\cite{lin2014microsoft} contains approximately 82K training images and 40504 validation images containing objects of 80 classes. As to the performance metric, we adopt the mean Intersection-over-Union (mIoU) between ground truth and predicted segmentation results. The results for the test images are obtained from the official PASCAL VOC evaluation website.
	
	\subsection{Implementation Details}
	\label{sec3_2}
	
	In our proposed approach, ResNet50~\cite{he2016deep} is adopted as our classification network by replacing the final fully-connected layer with a full convolution layer. We embed patch learning method into the stage 2, stage 3 and stage 4 of ResNet50 architecture since the discriminative learning ability of neural network stage 1 is much less obvious as shown in Fig.~\ref{fig:fig_3}.
	We also change the stride in stage 4 from 2 to 1 to get larger spatial dimension. Thus, the width and height of the feature maps are $\frac{1}{16}$ of the input image. The backbone network is initialized with the pretrained weights from ImageNet~\cite{deng2009imagenet}. Denoting the step where the stage 2, stage 3 and stage 4 in ResNet50 are destructed as the $1^{th}$ step, and the step where only the stage 3 or stage 4 in ResNet50 is destructed as the $2^{nd}$ or $3^{rd}$ step, $C_{1,1}, C_{1,2}, C_{1,3}, C_{2,1}, C_{2,2}, C_{2,3}$ are set as 256, 512, 1024, 2048, 2048 and 2048, respectively. Our final chosen parameter of $K$ in Patch Operation and Merge Operation are 2, 4 and 6.
	
	We use horizontal flip, random rescale in the range of $\begin{bmatrix}320, 640\end{bmatrix}$ by the longest edge, and then crop and pad the raw image by the size $512\times512$ as the network input, which imposes scale invariance in the network training. All the added layers are initialized by Kaiming uniform distribution~\cite{he2015delving}. We adopt SGD optimizer with momentum 0.9 and weight decay $1e^{-4}$. The initial learning rate is set to 0.1 and followed the poly policy $lr_{iter}=lr_{init}(1-\frac{iter}{max\_iter})^\gamma$ and the learning rate for new layers is multiplied by 10. For generating CAM map, we adopt multiple scales fusion, \textit{e.g.}, $0.5, 1.0, 1.5, 2.0$ to get more invariant CAM results and sum them up as ~\cite{chang2020weakly,ahn2018learning}, and we compute the CAM map via our proposed CAM generation method in Sec~\ref{sec2_5} and adopt refinement method to expand them to obtain final pseudo semantic segmentation labels as mentioned in Sec~\ref{sec2_6}. {For PL training for the implicit method, we adopt 5 epochs for finetuning the classification network, total 3305 iterations for PASCAL VOC 2012 and total 25865 iterations for MS COCO 2014, respectively. Meanwhile, as we fix the previous layers while training proceeds, we use a smaller learning rate 0.01 to fine-tune our PL training for both PASCAL VOC 2012 and MS COCO 2014.} {For initial CAM seed refinement, we utilize the IRNet to get our refined pseudo segmentation labels. The displacement field branch receives are amplified by a factor of 10. The image input are cropped into (512, 512). Meanwhile, the training learning rate is set to 0.1 and training batch size is 32. The refined network IRNet are trained for 3 epochs. After that we apply 8 times for controlling the random walk to obtain the refined segmentation labels.}

	\begin{table}[]
		\renewcommand\arraystretch{1.5}
		\setlength{\tabcolsep}{1.5mm}
		\centering
		\captionsetup{font={small}}
		\caption{{CAM quality measured by mIoU (\%) based on different classification backbones. ``Seed'', ``CRF'' and ``Refinement'' indicates the results without refinement, with CRF postprocessing and with refinement, respectively.}}
		\begin{tabular}{lcccc}
			\toprule[1pt]
			Method               & Seed &             CRF &     Refinement&                       Classification\\
			\midrule
			\midrule
			\multicolumn{5}{l}{Seed Refine with AffiniyNet~\cite{ahn2018learning}:} \\
			AffinityNet~\cite{ahn2018learning}      & 48.9         & -         & 59.6             & VGG16       \\
			\rowcolor{gray!25} Ours (implicit)      & \textbf{49.7}   & -     & \textbf{61.8}    & VGG16       \\
			\midrule
			\multicolumn{5}{l}{Seed Refine with AffiniyNet~\cite{ahn2018learning}:} \\
			AffinityNet~\cite{ahn2018learning}      & 48.0            & -      & 59.7             & WideResNet-38    \\
			SC-CAM~\cite{chang2020weakly}           & 50.9            & -      & 63.4             & WideResNet-38    \\
			MixUp-CAM~\cite{chang2020mixup}      & 50.1       & -  & 61.9             & WideResNet-38    \\
			\midrule
			\multicolumn{5}{l}{Seed Refine with IRNet~\cite{ahn2019weakly}:} \\
			IRNet~\cite{ahn2019weakly}            & 48.4         & 53.7         & 66.3                   & ResNet50    \\
			BE~\cite{chen2020weakly}               & 49.6        & -         & 67.2                   & ResNet50    \\
			CONTA~\cite{zhang2020causal}               & 48.8    & -   & 67.9                   & ResNet50    \\
			MBMNet~\cite{liu2020weakly}               & 50.2     & -   & 66.8                   & ResNet50    \\
			\rowcolor{gray!25} Ours (implicit)      & \textbf{50.6}   & \textbf{56.90}     & \textbf{67.7}          & ResNet50    \\
			\rowcolor{gray!25} Ours (explicit)      & \textbf{51.0}   & \textbf{57.51}      & \textbf{68.5}          & ResNet50    \\
			\bottomrule[1pt]
		\end{tabular}
		\label{tab:Table_1}
	\end{table}
	
	\begin{table}
		\centering
		\setlength{\tabcolsep}{1.5mm}
		\caption{\small Comparison with related works on the PASCAL VOC 2012 \textit{val} and \textit{test} set. The training supervision (Sup.) indicates: $\mathcal{I}$-image-level label, $\mathcal{B}$-bounding box, $\mathcal{S}$-scribble, and $\mathcal{F}$-segmentation label. ``Extra Info'' means training with additional information, ``Sal'' adopts saliency model and ``Sps'' adopts pre-computed superpixels. In addition, we present methods that aim to improve the initial CAM map quality with $\surd$ in the ``Init.'' column.}
		\begin{tabular}{l|cccccc}
			\toprule
			Method & \multicolumn{1}{c}{Sup.} & \multicolumn{1}{c}{Backbone} & \multicolumn{1}{c}{Extra Info} & \multicolumn{1}{c}{Init.} & \multicolumn{1}{c}{val} & \multicolumn{1}{c}{test}   \\
			\midrule
			WSSL~\cite{papandreou2015weakly}          & $\mathcal{B}$     & VGG-16             &                            &                         &  60.6                   & 62.2 \\
			BoxSup~\cite{dai2015boxsup}               & $\mathcal{B}$     & VGG-16             &                            &                         &  62.0                   & 64.6  \\
			SDI~\cite{khoreva2017simple}              & $\mathcal{B}$     & VGG-16             & Extra data                 &                         &  65.7                   & 67.5 \\
			\midrule
			Scribblesup~\cite{lin2016scribblesup}     & $\mathcal{S}$     & VGG-16             &                            &                         &  63.1                   & - \\
			PSCL~\cite{ke2021universal}               & $\mathcal{S}$       & ResNet-101         &                            &                       & 76.1                    & 76.4 \\
			BPG~\cite{wang2019boundary}               & $\mathcal{S}$       & ResNet-101         &                          &                        & 76.0                   & -  \\
			\midrule
			FCN~\cite{long2015fully}                  & $\mathcal{F}$     & VGG-16             &                            &                         &  -                      & 62.2 \\
			DeepLab~\cite{chen2017deeplab}            & $\mathcal{F}$     & VGG-16             &                            &                         &  67.6                   & 70.3 \\
			\midrule
			SEC~\cite{kolesnikov2016seed}             & $\mathcal{I}$     & VGG-16             &                            &                         & 50.7                    & 51.1 \\
			STC~\cite{wei2016stc}                     & $\mathcal{I}$     & VGG-16             &  Sal                       &                         & 49.8                    & 51.2 \\
			AE-PSL~\cite{wei2017object}               & $\mathcal{I}$     & VGG-16             &  Sal                       &                         & 55.0                    & 55.7 \\
			MDC~\cite{wei2018revisiting}              & $\mathcal{I}$     & VGG-16             &  Sal                       &                         & 60.4                    & 60.8 \\
			\midrule
			MCOF~\cite{wang2018weakly}                & $\mathcal{I}$     & ResNet-101         &  Sal                       &                         & 60.3                    & 61.2 \\
			SeeNet~\cite{hou2018self}                 & $\mathcal{I}$     & ResNet-101         &  Sal                       & $\surd$                 & 63.1                    & 62.8 \\
			DSRG~\cite{huang2018weakly}               & $\mathcal{I}$     & ResNet-101         &  Sal                       &                         & 61.4                    & 63.2 \\
			AffinityNet~\cite{ahn2018learning}        & $\mathcal{I}$     & ResNet-38          &                            &                         & 61.7                    & 63.7 \\
			FickleNet~\cite{lee2019ficklenet}         & $\mathcal{I}$     & ResNet-101         &  Sal                       & $\surd$                 & 64.9                    & 65.3 \\
			OAA~\cite{jiang2019integral}              & $\mathcal{I}$     & ResNet-101         &  Sal                       & $\surd$                 & 63.9                    & 65.6 \\
			CIAN~\cite{fan2020cian}                   & $\mathcal{I}$     & ResNet-101         &  Sal                       &                         & 64.1                    & 64.7 \\
			SC-CAM~\cite{chang2020weakly}             & $\mathcal{I}$     & ResNet-101         &                            & $\surd$                 & 66.1                    & 65.9 \\
			ICD~\cite{fan2020learning}                & $\mathcal{I}$     & ResNet-101         &  Sps                       &                         & \textbf{67.8}           & 68.0 \\
			SEAM~\cite{wang2020self}                  & $\mathcal{I}$     & ResNet-38          &                            & $\surd$                 & 64.5                    & 65.7 \\
			SvM~\cite{zhang2020splitting}             & $\mathcal{I}$     & ResNet-50          &  Sal                       & $\surd$                 & 66.6                    & 66.7 \\
			EME~\cite{fanemploying}                   & $\mathcal{I}$     & ResNet-101         &  Sal                       &                         & 67.2                    & 66.7 \\
			MCI~\cite{sun2020mining}                  & $\mathcal{I}$     & ResNet-101         &                            & $\surd$                 & 66.2                    & 66.9 \\
			MixUp-CAM~\cite{chang2020mixup}           & $\mathcal{I}$     & ResNet-101         &                            & $\surd$                 & 65.6                    & - \\
			IRNet~\cite{ahn2019weakly}                & $\mathcal{I}$     & ResNet-50          &                            &                         & 63.5                    & 64.8 \\
			MBMNet~\cite{liu2020weakly}               & $\mathcal{I}$     & ResNet-101         &           
			&                         & 66.2                    & 67.1  \\
			CONTA~\cite{zhang2020causal}              & $\mathcal{I}$     & ResNet-101         &           
			&                         & 65.3                    & 66.1  \\
			BE~\cite{chen2020weakly}                  & $\mathcal{I}$     & ResNet-101         &                            &                         & 65.7                    & 66.6 \\
			LVW~\cite{ru2021learning}                & $\mathcal{I}$     & ResNet-101         &                            &                         & 67.2                    & 67.3 \\
			\midrule
			\rowcolor{gray!25} Ours {(implicit)}                                      &$\mathcal{I}$      & ResNet-101         &                            &  $\surd$                      & \textbf{67.5}                    & \textbf{69.4\footnotemark[1]} \\
			\rowcolor{gray!25} Ours {(explicit)}                                      &$\mathcal{I}$      & ResNet-101         &                            &  $\surd$                      & \textbf{67.8}                    & \textbf{69.6\footnotemark[2]} \\
			\bottomrule
		\end{tabular}
		\label{tab:comparisons}
	\end{table}
	\footnotetext[1]{http://host.robots.ox.ac.uk:8080/anonymous/V8UCCZ.html}
	\footnotetext[2]{http://host.robots.ox.ac.uk:8080/anonymous/1M5V0I.html}

	\subsection{Comparison With State-of-the-Art Methods}
	\label{sec3_3}
	
	\subsubsection{Comparison in terms of the CAM Quality}
\begin{figure}
		\centering
		\includegraphics[width=0.48\textwidth]{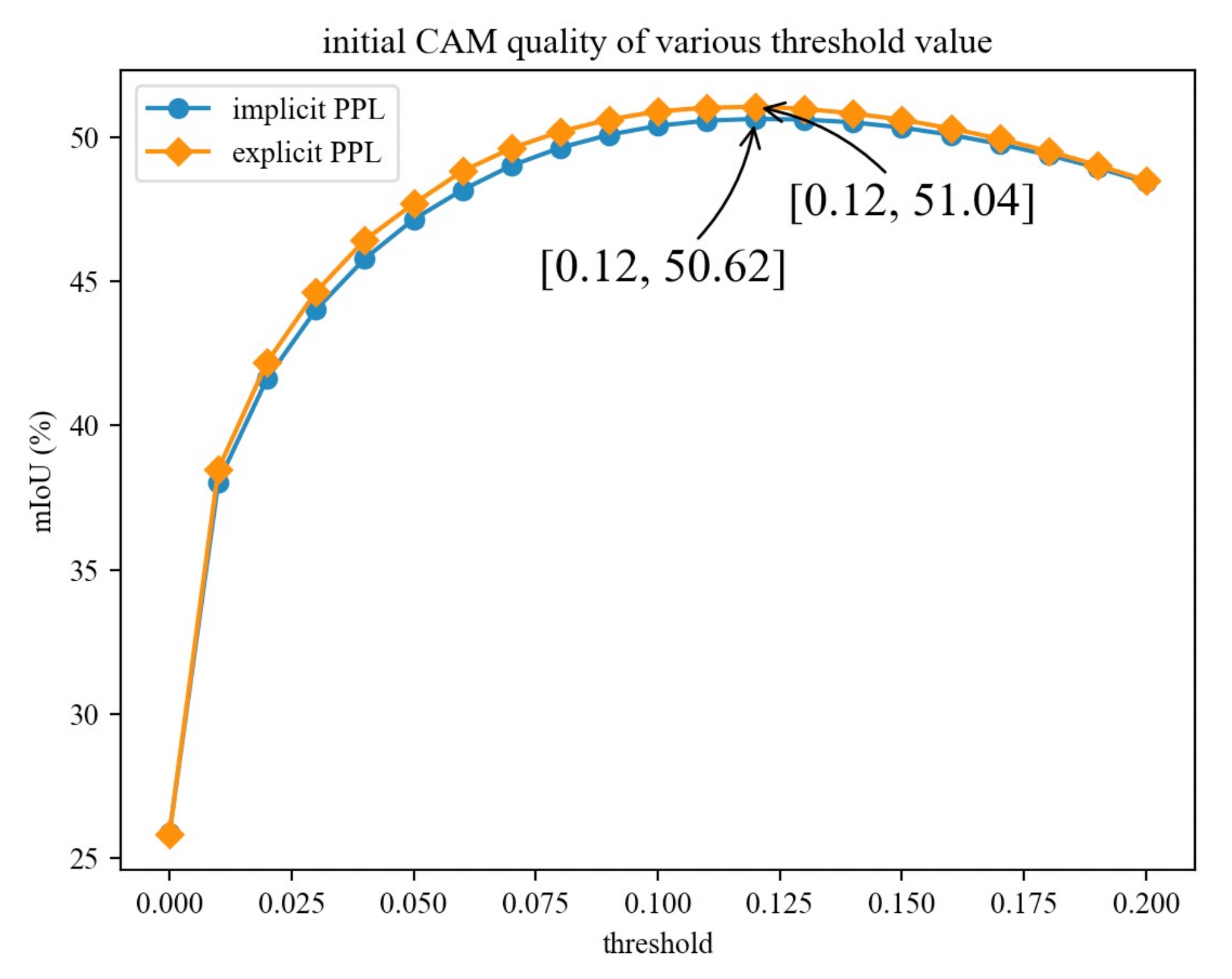}
		\captionsetup{font={small}}
		\caption{Initial CAM quality of various threshold values, in terms of mIoU (\%) on \textit{train} set. We also point out the best result on specific threshold value for the implicit and the explicit setting.}
		\label{fig:fig_cam}
	\end{figure}

	The proposed method is firstly compared with previous state-of-the-art image-level WSSS methods that adopt ResNet-50 as classification backbone and refinement procedure ~\cite{ahn2019weakly,chen2020weakly,liu2020weakly,zhang2020causal}. As shown in TABLE~\ref{tab:Table_1} bottom row, both our implicit and explicit method all outperforms most existing ResNet-50~\cite{he2016deep} classification based methods while compared with initial CAM seed or following refinement procedure in terms of mIoU of CAM quality on VOC12 \textit{train} set. Besides, we make comparisons with Wider-ResNet38~\cite{wu2019wider} classification based methods in TABLE~\ref{tab:Table_1} middle row, our methods are still superior, noting that Wider-ResNet38 are more powerful than ResNet-50. We also conduct experiment on VGG-16~\cite{simonyan2014very} classification based method AffinityNet~\cite{ahn2018learning} in TABLE~\ref{tab:Table_1} top row. Training details are almost the same as AffinityNet~\cite{ahn2018learning}, we replace the IRNet (ResNet-50) with AffinityNet (VGG-16) as second refinement method to be fair with AffinityNet. We obtain \textbf{49.7\%} mIoU and \textbf{61.8\%} mIoU on \textit{train} set \textit{vs.} AffinityNet~\cite{ahn2018learning} 48.9\% mIoU and 59.6\% mIoU for initial CAM quality and refined pseudo label mask quality, respectively. Since VGG-16 is much shallower than ResNet-50, and the destructed feature tiles are merged too quickly, thereby the progressive learning mechanism is less powerful than the ResNet-50 based. All these validate the effectiveness of our proposed methods, including the implicit and the explicit. Moreover, we draw a precision and recall curve compared with some SOTAs as shown in Fig.~\ref{fig:fig_pr}. We can see that our precision and recall curves stay higher than others which demonstrates the superiority of our methods, including both the implicit and the explicit fusion settings. For evaluating the impact of various threshold values on initial CAM seed quality, we conduct different threshold values to obtain various CAM seed quality and obtain their mIoU performance as shown in Fig.~\ref{fig:fig_cam}. The results show that our generated CAM seed quality is stable from 0.1 to 0.3 threshold values range, which demonstrates the CAM seed quality stability of our implicit and explicit method.

	\begin{table*}[htbp]
		\footnotesize
		\centering
		\setlength{\tabcolsep}{0.9mm}
		\caption{Evaluation on the PASCAL VOC 2012 \textit{val} and \textit{test} set.}
		\begin{tabular}{c|ccccccccccccccccccccc|c}
			\toprule
			Method & \multicolumn{1}{c}{bkg} & \multicolumn{1}{c}{aero} & \multicolumn{1}{c}{bike} & \multicolumn{1}{c}{bird} & \multicolumn{1}{c}{boat} & \multicolumn{1}{c}{bottle} & \multicolumn{1}{c}{bus} & \multicolumn{1}{c}{car} & \multicolumn{1}{c}{cat} & \multicolumn{1}{c}{chair} & \multicolumn{1}{c}{cow} & \multicolumn{1}{c}{table} & \multicolumn{1}{c}{dog} & \multicolumn{1}{c}{horse} & \multicolumn{1}{c}{mbk} & \multicolumn{1}{c}{person} & \multicolumn{1}{c}{plant} & \multicolumn{1}{c}{sheep} & \multicolumn{1}{c}{sofa} & \multicolumn{1}{c}{train} & \multicolumn{1}{c}{tv} & \multicolumn{1}{c}{mean} \\
			\midrule
			\midrule
			Ours implicit (\textit{val})  & 88.9  & 71.3  & 68.7  & 76.3  & 60.4  & 64.3  & 87.8  & 77.0  & 80.3  & 35.6  & 74.0  & 37.0  & 73.4  & 72.8  & 77.0  & 72.5  & 49.8  & 74.6  & 46.7  & 69.7  & 58.8  & 67.5  \\
			Ours explicit (\textit{val})  & 90.0  & 71.0  & 69.3  & 81.2  & 62.2  & 63.1  & 88.9  & 75.9  & 78.6  & 40.6  & 71.5 & 38 & 71.5  & 70.7  & 74.1  & 72.5  & 49.3  & 75.1  & 54.1  & 68.7  & 57.9  & 67.8  \\
			\midrule
			Ours implicit (\textit{test}) & 90.8  & 79.2  & 32.7  & 87.2  & 56.4  & 63.0  & 88.8  & 82.6  & 85.2  & 32.5  & 80.2  & 43.8  & 79.6  & 80.8  & 81.0  & 75.0  & 50.5  & 82.8  & 61.1  & 65.0  & 58.5  & 69.4  \\
			Ours explicit (\textit{test}) & 91.0  & 74.3  & 32.0  & 86.3  & 54.9  & 67.5  & 88.1  & 80.9  & 87.9  & 31.7  & 78.8  & 55.1  & 85.0  & 80.7  & 81.6  & 72.8  & 47.4  & 84.6  & 55.3  & 67.2  & 58.1  & 69.6  \\
			\bottomrule
		\end{tabular}%
		\label{tab:comparisons_class_test}
	\end{table*}%

	\subsubsection{Comparison in terms of Segmentation Performance}
	To further evaluate the performance of pseudo pixel-level annotations, we follow the common practise~\cite{chen2020weakly, lee2019ficklenet,fan2020learning,chang2020weakly} to train a DeepLab V2~~\cite{chen2017deeplab} segmentation model with ResNet-101 architecture, with full supervision of our generated pseudo segmentation masks, including the implicit and the explicit one. In addition, we apply CRF~\cite{krahenbuhl2011efficient} post-processing to further refine the final segmentation results as common practice. TABLE~\ref{tab:comparisons} shows that our implicit version approach already outperforms other state-of-the-art methods on VOC 2012 \textit{test} set utilizing only image-level labels, achieving \textbf{69.4\%} mIoU value, and the explicit version further improves the result to \textbf{69.6\%}. For VOC 2012 \textit{val} set, our proposed explicit method achieves competitive mIoU performance with ICD~\cite{fan2020learning} adopting external pre-computed superpixels information, and surpasses EME~\cite{fanemploying} model, which applies multiple CAM seeds fusion strategies to train the segmentation network. Our approach (implicit) surpasses IRNet~\cite{ahn2019weakly}, which applies the same refinement method as ours, by \textbf{4.0\%} and \textbf{4.6\%} on \textit{val} and \textit{test} set, respectively. For the explicit fusion setting, our approach (explicit) surpasses IRNet by \textbf{4.3\%} and \textbf{4.8\%} on \textit{val} and \textit{test} set, respectively. Thanks to higher quality initial CAM seeds provided by our method, our approach still outperforms other methods introducing extra training data SDI~\cite{khoreva2017simple} and most others adopting saliency model refinement, which need another pixel-level annotation saliency benchmark to train an offline salient model. Examples of semantic segmentation results provided in Fig.~\ref{fig:vis_seg} show that our segmentation results are close to the ground truth, even in handling multiple instances image scene. We also present our per-class IoU performance in TABLE~\ref{tab:comparisons_class_test}, the result of \textit{test} set are obtained by submitting the \textit{test} results to PASCAL VOC 2012 server.

	{To further validate the competitive performance of our methods, we conduct experiments on MS COCO2014 dataset~\cite{lin2014microsoft} as shown in TABLE~\ref{tab:coco_comparison}. Our implicit results still outperform the most existing works. Our implicit method on MS-COCO obtains \textbf{29.1\%}, \textbf{35.2\%} and \textbf{33.4\%} mIoU for initial CAM Seed, refined pseudo mask and final segmentation, respectively. Our explicit method on MS-COCO achieves \textbf{29.8\%}, \textbf{36.1\%} and \textbf{34.4\%} mIoU for initial CAM Seed, refined pseudo mask and final segmentation, respectively.}

	\begin{table}
		\renewcommand\arraystretch{1.5}
		\setlength{\tabcolsep}{1.5mm}
		\centering
		\captionsetup{font={small}}
		\caption{{CAM quality measured by mIoU (\%) based on different classification backbones on MS COCO 2014. ``Seed'',  and ``Refinement'' indicates the results without and with refinement, respectively.}}
		\begin{tabular}{lcccc}
			\toprule[1pt]
			Method                         & Seed   & Refinement  & Segmentation  & Classification\\
			\midrule
			IRNet~\cite{ahn2019weakly}     & 27.4   & 34.0        & 32.6           & ResNet50    \\
			SEAM~\cite{wang2020self}       & 25.1   & 31.5        & 32.8           & Wide-ResNet38    \\
			CONTA~\cite{zhang2020causal}   & 28.7   & 35.2        & 33.4           & ResNet50    \\
			\rowcolor{gray!25} Ours (implicit)                  & \textbf{29.1}   & \textbf{35.7}        & \textbf{33.8}          & ResNet50    \\
			\rowcolor{gray!25} Ours (explicit)                  & \textbf{29.8}   & \textbf{36.1}        & \textbf{34.4}           & ResNet50    \\
			\bottomrule[1pt]
		\end{tabular}
		
		\label{tab:coco_comparison}
	\end{table}
	
	\subsection{Analysis of Pseudo Labels}
	\label{sec3_4}
	
	\subsubsection{Comparison with the Baseline}
	Our proposed framework is compared with the baseline pseudo segmentation label synthesis models, CAM~\cite{zhou2016learning} and CAM+IRNet~\cite{ahn2019weakly}, in terms of mIoU evaluated on PASCAL VOC 2012 \emph{train} set. As shown in TABLE~\ref{tab:Table_2}, our proposed framework results in \textbf{2.22\%}, \textbf{2.10\%} and \textbf{2.64\%}, \textbf{2.80\%} mIoU value improvement for the implicit and the explicit method, compared to the baseline models CAM and CAM+IRNet, respectively. {The visualization examples in Fig.~\ref{fig:vis_cam} show that our proposed framework is able to cover more complete object regions and closer to the ground truth mask not only for the samples that have single target object but also for the ones that have multiple target objects, while the original CAM only locates the discriminative object parts.}

	\begin{table}[]
		\centering
		\caption{Quality of CAM seeds for our proposed PPL method (implicit and explicit). CAM means initial CAM seeds obtained from the classifier. ``PL'' illustrates the proposed Patch Learning. ``PL S $i$'' means applying Patch Learning with specific $K$ patch operation before the $i^{th}$ ResNet Stage block. CAM+IRNet means the pseudo segmentation label propagation using the IRNet to refine, in terms of mIoU (\%) evaluated on the PASCAL VOC 2012 \textit{train} set.}
		\begin{tabular}{l|ccc}
			\toprule
			Setting                 & $K$    & CAM &                  CAM+IRNet                     \\
			\midrule
			\midrule
			baseline                & -      & 48.40                  & 65.30                       \\
			PL S 2                  & 2      & 49.74                  & 66.77                       \\
			PL S 3                  & 4      & 49.88                  & 66.97                       \\
			PL S 4                  & 6      & 49.87                  & 67.09                       \\
			\rowcolor{gray!25} Ours (implicit)         & -      & \textbf{50.62}         & \textbf{67.40}              \\
			\rowcolor{gray!25} Ours (explicit)         & -      & \textbf{51.04}         & \textbf{68.10}              \\
			\bottomrule
		\end{tabular}%
		\label{tab:Table_2}
	\end{table}
	
	\begin{table}[]
		\setlength{\tabcolsep}{1.3mm}
		\centering
		\caption{Quality of CAM seeds for Patch Learning with different grid size $K$ in the beginning of different ResNet block, in terms of mIoU (\%) evaluated on the PASCAL VOC 2012 \textit{train} set.}
		\begin{tabular}{cc||cc||cc}
			\toprule
			Stage 2 $K$         & {mIoU (\%)}& Stage 3 $K$         & {mIoU (\%)}&           Stage 4 $K$         & mIoU (\%)  \\
			\midrule
			\midrule
			2                  & \textbf{49.74}        & 2                  & 49.78                  & 2                  & 49.79               \\
			3                  & 49.30                 & 3                  & 49.15                  & 3                  & 49.63               \\
			4                  & 49.38                 & 4                  & \textbf{49.88}         & 4                  & 49.78               \\
			5                  & 48.91                 & 5                  & 49.07                  & 5                  & 49.32      \\
			6                  & 49.10                 & 6                  & 49.07                  & 6                  & \textbf{49.87}      \\
			7                  & 48.96                 & 7                  & 48.48                  & 7                  & 49.30               \\
			8                  & 49.28                 & 8                  & 49.29                  & 8                  & 49.05               \\
			\bottomrule
		\end{tabular}%
		
		\label{tab:Table_3}
	\end{table}

	\subsubsection{Ablation Studies}
	\label{sec3_5}
	
	To demonstrate the effectiveness of our approach, we conduct ablation studies on the PASCAL VOC 2012 \textit{train} set.
	
	\textbf{Patch Learning Effect.} 
	We conduct different patch learning operation at the beginning of different ResNet blocks with various $K$ values, Stage 2, Stage 3 and Stage 4. Specially, we embed Patch Operation before ResNet Stage 2, Stage 3 or Stage 4 to destruct the global feature structure into different tiles, and apply Merge Operation before the final classifier layer to obtain the concatenated feature maps. Noting that all the following convolution blocks are utilized before the final classifier layer. As shown in Table~\ref{tab:Table_3}, while embedding patch learning into different ResNet Stage block, we achieve \textbf{49.74\%} mIoU for grid size ($2 \times 2$) before ResNet Stage 2, \textbf{49.88\%} mIoU for grid size ($4 \times 4$) before ResNet Stage 3 and \textbf{49.87\%} before ResNet Stage 4 for grid size ($6 \times 6$) with remarkable improvements, compared to  48.40\% of the baseline. We further find out that using an appropriate Patch Learning grid size can lead to the optimal result while a certain range grid size also boosts the performance. We only conduct single Patch Learning method before a specific convolution block and obtain remarkable improvement thanks to the extra finer granularity local feature mining. And the visualization demos are shown in Fig.~\ref{fig:vis_cam} at column (c), (d), (e). The body of the bird in Fig.~\ref{fig:vis_cam} row two is re-activated by Patch Learning while it is ignored by the baseline model.
	
	\begin{figure}
		\centering
		\includegraphics[width=0.48\textwidth]{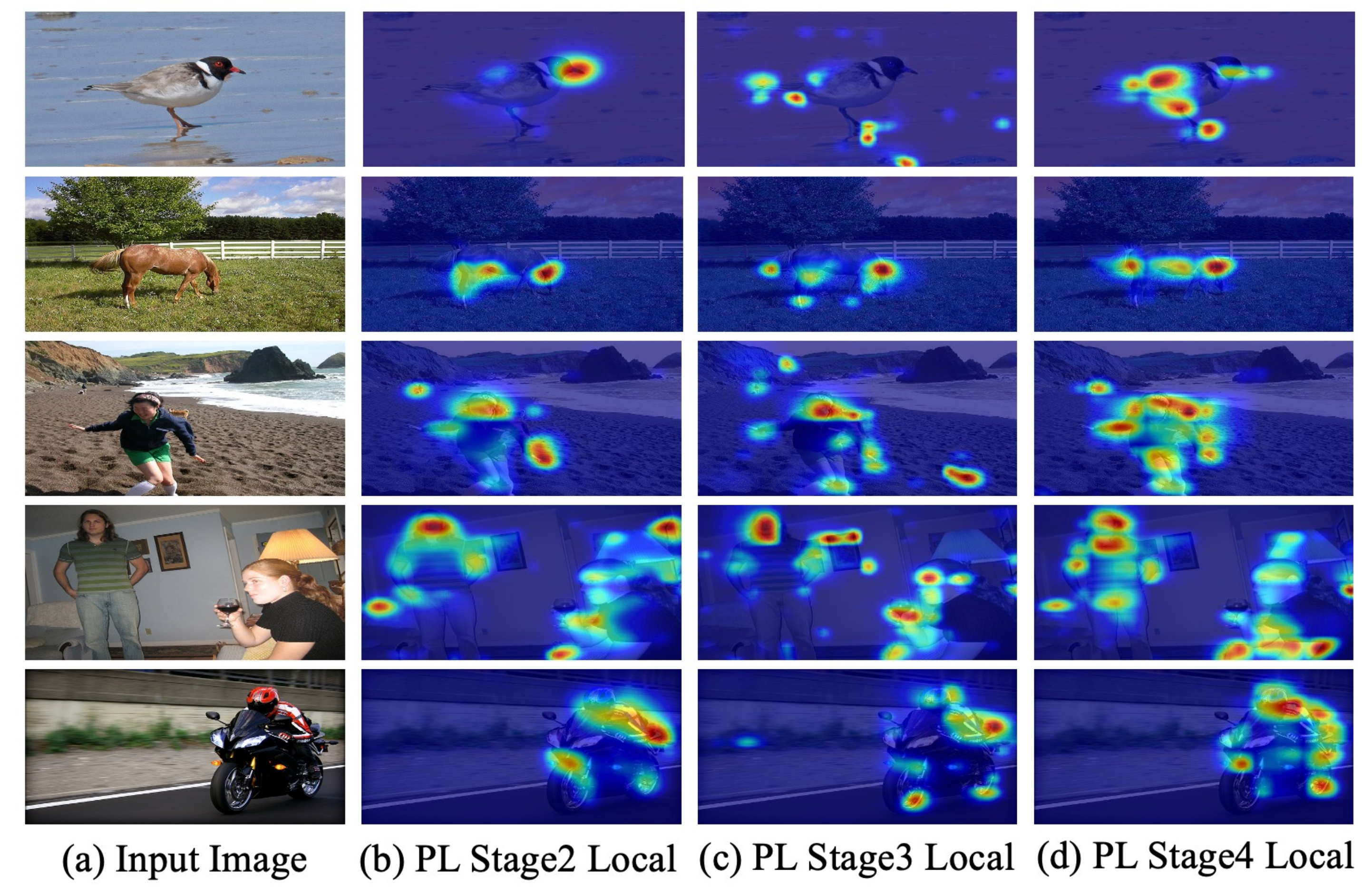}
		\captionsetup{font={small}}
		\caption{Visualization of the proposed Patch Learning that embedded into the ResNet Stage 2, 3 or 4 only and trained without concatenation with the global feature maps.}
		\label{fig:vis_cam_local}
	\end{figure}
	
	\begin{table}[]
		\centering
		\caption{Quality of CAM seeds for progressive training with different grid size combination between Stage 2, Stage 3 or Stage 4 in ResNet50 denoted as $i\_j$ or $i$\_$j\_k$ respectively in stage ordering, in terms of mIoU (\%) evaluated on the PASCAL VOC 2012 \textit{train} set.}
		\begin{tabular}{cc||cc}
			\toprule
			Setting $K$        & {mIoU (\%)}&           Setting $K$        & mIoU (\%)  \\
			\midrule
			\midrule
			2\_2                  & 49.99                  & 4\_2                  & 50.25               \\
			2\_4                  & 50.07                  & 4\_4                  & 50.18               \\
			2\_6                  & \textbf{50.22}         & 4\_6                  & \textbf{50.42}               \\
			2\_8                  & 49.64                  & 4\_8                  & 50.06      \\
			\midrule
			\midrule
			2\_2\_6                  & 49.95                  & 2\_4\_6                  & \textbf{50.62}                \\
			4\_2\_6                  & 50.01                  & 4\_4\_6                  & 50.38               \\
			6\_2\_6                  & 50.00         & 6\_4\_6                           & 50.30               \\
			8\_2\_6                  & 49.88                  & 8\_4\_6                  & 50.06      \\
			\bottomrule
		\end{tabular}%
		\label{tab:Table_4}
	\end{table}
	
	\textbf{Implicit Fusion by Progressive Training.} 
	As TABLE~\ref{tab:Table_4} shows, we obtain the best performance \textbf{50.62\%} mIoU for the combination that trains network embedding Patch Learning with grid size ($2 \times 2$) before Stage 2, and progressively moves to apply grid size ($4 \times 4$) before Stage 3, and then fixes the layers before Stage 4 and trains on Stage 4. Noting that we achieve this via progressive training strategy. Fig.~\ref{fig:vis_cam} illustrates the optimal CAM results are obtained while applying Patch Learning before ResNet Stage 2 ($K=2$), Stage 3 ($K=4$) and Stage 4 ($K=6$) breaking up features in one manner. We then move further to conduct a progressive training strategy. At TABLE~\ref{tab:Table_3}, the final denoted combination ``$2\_4\_6$'' is reached, firstly trains the network applying Patch Learning before Stage 2, then moves to the top layer until Stage 4. We analyze that the bottom layer is able to capture finer feature granularities while the top layer becomes coarse. Therefore, a larger grid size is required during destruction when coming to higher layers. This guarantees that the network is forced to potentially aggregate more non-discriminative cues to optimize the final classification task.

	\textbf{Without Global Feature.}
	We also show our results without concatenation with the global feature $\mathbf{\hat{F}}$. In this case, only the collected feature $\tilde{\mathbf{F}}$ is fed into the final classification layer. As shown at the top row of TABLE~\ref{tab:Table_5}, we conduct experiments on single Patch Learning before Stage 2 ($K=2$), Stage 3 ($K=4$) and Stage 4 ($K=6$), then we get 26.30\%, 16.17\% and 24.92\% mIoU, respectively. Some examples are visualized in Fig.~\ref{fig:vis_cam_local}. We can find that Patch Learning method enforces the network to seek other weak local information without concatenation with global image feature. For example, the legs of horse or the body of the bird is activated by patch learning while it is usually ignored by the common classification model.
	
	\begin{table}[]
		\centering
		\caption{Quality of CAM seeds for explicit concatenation with different grid size combination between Stage 2, Stage 3 or Stage 4 in ResNet50 denoted as $i$\_$j\_k$ respectively, in terms of mIoU (\%) evaluated on the PASCAL VOC 2012 \textit{train} set.}
		\begin{tabular}{l|lc}
			\toprule
			Setting $K$         & concatenation                                    & {mIoU (\%)}     \\
			\midrule
			\midrule
			6               & PL S 4                  & 26.30             \\
			4               & PL S 3                  & 16.17             \\
			2               & PL S 2                  & 24.92            \\
			\midrule
			\midrule
			6               & PL S 4 + global                          & 49.87             \\
			4\_6            & PL S 3 + PL S 4 + global                 & 50.78             \\
			2\_4\_6         & PL S 2 + PL S 3 + PL S 4 + global        & \textbf{51.04}   \\
			2\_2\_4\_6      & PL S 1 + PL S 2 + PL S 3 + PL S 4 + global     & 51.01   \\
			\bottomrule
		\end{tabular}%
		\label{tab:Table_5}
	\end{table}
	
	\textbf{Explicit Fusion.}
	We conduct experiments to validate the effectiveness of the explicit fusion approach. With multi-branch shared network training, we get \textbf{51.04\%} mIoU for the initial CAM result at the bottom of TABLE~\ref{tab:Table_5}. Then we utilize IRNet~\cite{ahn2019weakly} to refine pseudo segmentation label and train the segmentation model. As shown in Table~\ref{tab:comparisons}, the mIoU of WSSS results reach 67.8\% for \textit{val} and 69.6\% for \textit{test} compared to 67.5\% for \textit{val} and 69.4\% for \textit{test} of the implicit version. It demonstrates that both the implicit progressive fusion and explicit fusion boost final WSSS performance. However the explicit strategy inevitably costs more GPU memory resource than the implicit one. We also visualize some initial CAM map in Fig.~\ref{fig:vis_cam} (g column) for the \textit{train} set and final semantic segmentation results on the \textit{val} set in Fig.~\ref{fig:vis_seg}.
	
	\section{Conclusion}
	In this paper, we propose a novel Progressive Patch Learning method to improve the quality of CAMs from bottom to top layer feature enhancement. Specifically, Patch Learning destructs feature maps into patches and forces the network to seek  weaker cues from scattered discriminative local parts. We further apply such destruction and patch learning to multi-level granularities in a progressive manner. Moreover, we propose an explicit fusion method to gather multi-level granularities simultaneously with less training time. Thus, the network is trained to accumulate diverse feature response granularities among different convolution blocks. Consequently, the classification network has more chance to cover the whole object regions achieving outstanding semantic segmentation performance.
	
	\ifCLASSOPTIONcaptionsoff
	\newpage
	\fi
	
	
	
	%
	
	\bibliographystyle{IEEEtran}
	\bibliography{ieetrans}

\begin{thebibliography}{10}
\providecommand{\url}[1]{#1}
\csname url@samestyle\endcsname
\providecommand{\newblock}{\relax}
\providecommand{\bibinfo}[2]{#2}
\providecommand{\BIBentrySTDinterwordspacing}{\spaceskip=0pt\relax}
\providecommand{\BIBentryALTinterwordstretchfactor}{4}
\providecommand{\BIBentryALTinterwordspacing}{\spaceskip=\fontdimen2\font plus
\BIBentryALTinterwordstretchfactor\fontdimen3\font minus
  \fontdimen4\font\relax}
\providecommand{\BIBforeignlanguage}[2]{{%
\expandafter\ifx\csname l@#1\endcsname\relax
\typeout{** WARNING: IEEEtran.bst: No hyphenation pattern has been}%
\typeout{** loaded for the language `#1'. Using the pattern for}%
\typeout{** the default language instead.}%
\else
\language=\csname l@#1\endcsname
\fi
#2}}
\providecommand{\BIBdecl}{\relax}
\BIBdecl

\bibitem{chen2017deeplab}
L.-C. Chen, G.~Papandreou, I.~Kokkinos, K.~Murphy, and A.~L. Yuille, ``Deeplab:
  Semantic image segmentation with deep convolutional nets, atrous convolution,
  and fully connected crfs,'' \emph{IEEE transactions on pattern analysis and
  machine intelligence}, vol.~40, no.~4, pp. 834--848, 2017.

\bibitem{chen2014semantic}
------, ``Semantic image segmentation with deep convolutional nets and fully
  connected crfs,'' in \emph{Proceedings of the International Conference on
  Learning Representations}, 2015.

\bibitem{long2015fully}
J.~Long, E.~Shelhamer, and T.~Darrell, ``Fully convolutional networks for
  semantic segmentation,'' in \emph{Proceedings of the IEEE conference on
  computer vision and pattern recognition}, 2015, pp. 3431--3440.

\bibitem{wei2018revisiting}
Y.~Wei, H.~Xiao, H.~Shi, Z.~Jie, J.~Feng, and T.~S. Huang, ``Revisiting dilated
  convolution: A simple approach for weakly-and semi-supervised semantic
  segmentation,'' in \emph{Proceedings of the IEEE Conference on Computer
  Vision and Pattern Recognition}, 2018, pp. 7268--7277.

\bibitem{ahn2018learning}
J.~Ahn and S.~Kwak, ``Learning pixel-level semantic affinity with image-level
  supervision for weakly supervised semantic segmentation,'' in
  \emph{Proceedings of the IEEE Conference on Computer Vision and Pattern
  Recognition}, 2018, pp. 4981--4990.

\bibitem{wei2017object}
Y.~Wei, J.~Feng, X.~Liang, M.-M. Cheng, Y.~Zhao, and S.~Yan, ``Object region
  mining with adversarial erasing: A simple classification to semantic
  segmentation approach,'' in \emph{Proceedings of the IEEE conference on
  computer vision and pattern recognition}, 2017, pp. 1568--1576.

\bibitem{wang2018weakly}
X.~Wang, S.~You, X.~Li, and H.~Ma, ``Weakly-supervised semantic segmentation by
  iteratively mining common object features,'' in \emph{Proceedings of the IEEE
  conference on computer vision and pattern recognition}, 2018, pp. 1354--1362.

\bibitem{kolesnikov2016seed}
A.~Kolesnikov and C.~H. Lampert, ``Seed, expand and constrain: Three principles
  for weakly-supervised image segmentation,'' in \emph{European conference on
  computer vision}.\hskip 1em plus 0.5em minus 0.4em\relax Springer, 2016, pp.
  695--711.

\bibitem{huang2018weakly}
Z.~Huang, X.~Wang, J.~Wang, W.~Liu, and J.~Wang, ``Weakly-supervised semantic
  segmentation network with deep seeded region growing,'' in \emph{Proceedings
  of the IEEE Conference on Computer Vision and Pattern Recognition}, 2018, pp.
  7014--7023.

\bibitem{lin2016scribblesup}
D.~Lin, J.~Dai, J.~Jia, K.~He, and J.~Sun, ``Scribblesup: Scribble-supervised
  convolutional networks for semantic segmentation,'' in \emph{Proceedings of
  the IEEE Conference on Computer Vision and Pattern Recognition}, 2016, pp.
  3159--3167.

\bibitem{vernaza2017learning}
P.~Vernaza and M.~Chandraker, ``Learning random-walk label propagation for
  weakly-supervised semantic segmentation,'' in \emph{Proceedings of the IEEE
  conference on computer vision and pattern recognition}, 2017, pp. 7158--7166.

\bibitem{bearman2016s}
A.~Bearman, O.~Russakovsky, V.~Ferrari, and L.~Fei-Fei, ``What’s the point:
  Semantic segmentation with point supervision,'' in \emph{European conference
  on computer vision}.\hskip 1em plus 0.5em minus 0.4em\relax Springer, 2016,
  pp. 549--565.

\bibitem{dai2015boxsup}
J.~Dai, K.~He, and J.~Sun, ``Boxsup: Exploiting bounding boxes to supervise
  convolutional networks for semantic segmentation,'' in \emph{Proceedings of
  the IEEE international conference on computer vision}, 2015, pp. 1635--1643.

\bibitem{khoreva2017simple}
A.~Khoreva, R.~Benenson, J.~Hosang, M.~Hein, and B.~Schiele, ``Simple does it:
  Weakly supervised instance and semantic segmentation,'' in \emph{Proceedings
  of the IEEE conference on computer vision and pattern recognition}, 2017, pp.
  876--885.

\bibitem{papandreou1502weakly}
G.~Papandreou, L.-C. Chen, K.~Murphy, and A.~Yuille, ``Weakly-and
  semi-supervised learning of a deep convolutional network for semantic image
  segmentation,'' in \emph{Proceedings of the IEEE international conference on
  computer vision}, 2015, pp. 1742--1750.

\bibitem{zhou2016learning}
B.~Zhou, A.~Khosla, A.~Lapedriza, A.~Oliva, and A.~Torralba, ``Learning deep
  features for discriminative localization,'' in \emph{Proceedings of the IEEE
  conference on computer vision and pattern recognition}, 2016, pp. 2921--2929.

\bibitem{singh2017hide}
K.~K. Singh and Y.~J. Lee, ``Hide-and-seek: Forcing a network to be meticulous
  for weakly-supervised object and action localization,'' in \emph{2017 IEEE
  international conference on computer vision (ICCV)}.\hskip 1em plus 0.5em
  minus 0.4em\relax IEEE, 2017, pp. 3544--3553.

\bibitem{lee2019ficklenet}
J.~Lee, E.~Kim, S.~Lee, J.~Lee, and S.~Yoon, ``Ficklenet: Weakly and
  semi-supervised semantic image segmentation using stochastic inference,'' in
  \emph{Proceedings of the IEEE conference on computer vision and pattern
  recognition}, 2019, pp. 5267--5276.

\bibitem{jiang2019integral}
P.-T. Jiang, Q.~Hou, Y.~Cao, M.-M. Cheng, Y.~Wei, and H.-K. Xiong, ``Integral
  object mining via online attention accumulation,'' in \emph{Proceedings of
  the IEEE International Conference on Computer Vision}, 2019, pp. 2070--2079.

\bibitem{fan2020learning}
J.~Fan, Z.~Zhang, C.~Song, and T.~Tan, ``Learning integral objects with
  intra-class discriminator for weakly-supervised semantic segmentation,'' in
  \emph{Proceedings of the IEEE/CVF Conference on Computer Vision and Pattern
  Recognition}, 2020, pp. 4283--4292.

\bibitem{everingham2010pascal}
M.~Everingham, L.~Van~Gool, C.~K. Williams, J.~Winn, and A.~Zisserman, ``The
  pascal visual object classes (voc) challenge,'' \emph{International journal
  of computer vision}, vol.~88, no.~2, pp. 303--338, 2010.

\bibitem{selvaraju2017grad}
R.~R. Selvaraju, M.~Cogswell, A.~Das, R.~Vedantam, D.~Parikh, and D.~Batra,
  ``Grad-cam: Visual explanations from deep networks via gradient-based
  localization,'' in \emph{Proceedings of the IEEE international conference on
  computer vision}, 2017, pp. 618--626.

\bibitem{wang2020score}
H.~Wang, Z.~Wang, M.~Du, F.~Yang, Z.~Zhang, S.~Ding, P.~Mardziel, and X.~Hu,
  ``Score-cam: Score-weighted visual explanations for convolutional neural
  networks,'' in \emph{Proceedings of the IEEE/CVF conference on computer
  vision and pattern recognition workshops}, 2020, pp. 24--25.

\bibitem{zhang2021group}
Q.~Zhang, L.~Rao, and Y.~Yang, ``Group-cam: Group score-weighted visual
  explanations for deep convolutional networks,'' \emph{arXiv preprint
  arXiv:2103.13859}, 2021.

\bibitem{jiang2021layercam}
P.-T. Jiang, C.-B. Zhang, Q.~Hou, M.-M. Cheng, and Y.~Wei, ``Layercam:
  Exploring hierarchical class activation maps,'' \emph{IEEE Transactions on
  Image Processing}, 2021.

\bibitem{zhang2018top}
J.~Zhang, S.~A. Bargal, Z.~Lin, J.~Brandt, X.~Shen, and S.~Sclaroff, ``Top-down
  neural attention by excitation backprop,'' \emph{International Journal of
  Computer Vision}, vol. 126, no.~10, pp. 1084--1102, 2018.

\bibitem{zhang2013representative}
L.~Zhang, Y.~Gao, Y.~Xia, K.~Lu, J.~Shen, and R.~Ji, ``Representative discovery
  of structure cues for weakly-supervised image segmentation,'' \emph{IEEE
  transactions on multimedia}, vol.~16, no.~2, pp. 470--479, 2013.

\bibitem{zhang2019decoupled}
T.~Zhang, G.~Lin, J.~Cai, T.~Shen, C.~Shen, and A.~C. Kot, ``Decoupled spatial
  neural attention for weakly supervised semantic segmentation,'' \emph{IEEE
  Transactions on Multimedia}, vol.~21, no.~11, pp. 2930--2941, 2019.

\bibitem{hou2018self}
Q.~Hou, P.~Jiang, Y.~Wei, and M.-M. Cheng, ``Self-erasing network for integral
  object attention,'' in \emph{Advances in Neural Information Processing
  Systems}, 2018, pp. 549--559.

\bibitem{li2018tell}
K.~Li, Z.~Wu, K.-C. Peng, J.~Ernst, and Y.~Fu, ``Tell me where to look: Guided
  attention inference network,'' in \emph{Proceedings of the IEEE Conference on
  Computer Vision and Pattern Recognition}, 2018, pp. 9215--9223.

\bibitem{zhang2018adversarial}
X.~Zhang, Y.~Wei, J.~Feng, Y.~Yang, and T.~S. Huang, ``Adversarial
  complementary learning for weakly supervised object localization,'' in
  \emph{Proceedings of the IEEE Conference on Computer Vision and Pattern
  Recognition}, 2018, pp. 1325--1334.

\bibitem{wang2020self}
Y.~Wang, J.~Zhang, M.~Kan, S.~Shan, and X.~Chen, ``Self-supervised equivariant
  attention mechanism for weakly supervised semantic segmentation,'' in
  \emph{Proceedings of the IEEE/CVF Conference on Computer Vision and Pattern
  Recognition}, 2020, pp. 12\,275--12\,284.

\bibitem{chang2020weakly}
Y.-T. Chang, Q.~Wang, W.-C. Hung, R.~Piramuthu, Y.-H. Tsai, and M.-H. Yang,
  ``Weakly-supervised semantic segmentation via sub-category exploration,'' in
  \emph{Proceedings of the IEEE/CVF Conference on Computer Vision and Pattern
  Recognition}, 2020, pp. 8991--9000.

\bibitem{fan2020cian}
J.~Fan, Z.~Zhang, T.~Tan, C.~Song, and J.~Xiao, ``Cian: Cross-image affinity
  net for weakly supervised semantic segmentation,'' in \emph{Proceedings of
  the AAAI Conference on Artificial Intelligence}, vol.~34, no.~07, 2020, pp.
  10\,762--10\,769.

\bibitem{zhou2020sal}
L.~Zhou, C.~Gong, Z.~Liu, and K.~Fu, ``Sal: Selection and attention losses for
  weakly supervised semantic segmentation,'' \emph{IEEE Transactions on
  Multimedia}, vol.~23, pp. 1035--1048, 2020.

\bibitem{sun2020mining}
G.~Sun, W.~Wang, J.~Dai, and L.~Van~Gool, ``Mining cross-image semantics for
  weakly supervised semantic segmentation,'' in \emph{European Conference on
  Computer Vision}, 2020.

\bibitem{ru2021learning}
L.~Ru, B.~Du, and C.~Wu, ``Learning visual words for weakly-supervised semantic
  segmentation,'' in \emph{International Joint Conference on Artificial
  Intelligence}, 2021.

\bibitem{araslanov2020single}
N.~Araslanov and S.~Roth, ``Single-stage semantic segmentation from image
  labels,'' in \emph{Proceedings of the IEEE/CVF Conference on Computer Vision
  and Pattern Recognition}, 2020, pp. 4253--4262.

\bibitem{zhang2020reliability}
B.~Zhang, J.~Xiao, Y.~Wei, M.~Sun, and K.~Huang, ``Reliability does matter: An
  end-to-end weakly supervised semantic segmentation approach,'' in
  \emph{Proceedings of the AAAI Conference on Artificial Intelligence},
  vol.~34, no.~07, 2020, pp. 12\,765--12\,772.

\bibitem{shimoda2016distinct}
W.~Shimoda and K.~Yanai, ``Distinct class-specific saliency maps for weakly
  supervised semantic segmentation,'' in \emph{European Conference on Computer
  Vision}.\hskip 1em plus 0.5em minus 0.4em\relax Springer, 2016, pp. 218--234.

\bibitem{ahn2019weakly}
J.~Ahn, S.~Cho, and S.~Kwak, ``Weakly supervised learning of instance
  segmentation with inter-pixel relations,'' in \emph{Proceedings of the IEEE
  Conference on Computer Vision and Pattern Recognition}, 2019, pp. 2209--2218.

\bibitem{chen2020weakly}
L.~Chen, W.~Wu, C.~Fu, X.~Han, and Y.~Zhang, ``Weakly supervised semantic
  segmentation with boundary exploration.''\hskip 1em plus 0.5em minus
  0.4em\relax ECCV, 2020.

\bibitem{jiang2013salient}
H.~Jiang, J.~Wang, Z.~Yuan, Y.~Wu, N.~Zheng, and S.~Li, ``Salient object
  detection: A discriminative regional feature integration approach,'' in
  \emph{Proceedings of the IEEE conference on computer vision and pattern
  recognition}, 2013, pp. 2083--2090.

\bibitem{hou2017deeply}
Q.~Hou, M.-M. Cheng, X.~Hu, A.~Borji, Z.~Tu, and P.~H. Torr, ``Deeply
  supervised salient object detection with short connections,'' in
  \emph{Proceedings of the IEEE Conference on Computer Vision and Pattern
  Recognition}, 2017, pp. 3203--3212.

\bibitem{lee2021anti}
J.~Lee, E.~Kim, and S.~Yoon, ``Anti-adversarially manipulated attributions for
  weakly and semi-supervised semantic segmentation,'' in \emph{Proceedings of
  the IEEE/CVF Conference on Computer Vision and Pattern Recognition}, 2021,
  pp. 4071--4080.

\bibitem{wu2021embedded}
T.~Wu, J.~Huang, G.~Gao, X.~Wei, X.~Wei, X.~Luo, and C.~H. Liu, ``Embedded
  discriminative attention mechanism for weakly supervised semantic
  segmentation,'' in \emph{Proceedings of the IEEE/CVF Conference on Computer
  Vision and Pattern Recognition}, 2021, pp. 16\,765--16\,774.

\bibitem{noroozi2016unsupervised}
M.~Noroozi and P.~Favaro, ``Unsupervised learning of visual representations by
  solving jigsaw puzzles,'' in \emph{European Conference on Computer
  Vision}.\hskip 1em plus 0.5em minus 0.4em\relax Springer, 2016, pp. 69--84.

\bibitem{doersch2015unsupervised}
C.~Doersch, A.~Gupta, and A.~A. Efros, ``Unsupervised visual representation
  learning by context prediction,'' in \emph{Proceedings of the IEEE
  international conference on computer vision}, 2015, pp. 1422--1430.

\bibitem{noroozi2018boosting}
M.~Noroozi, A.~Vinjimoor, P.~Favaro, and H.~Pirsiavash, ``Boosting
  self-supervised learning via knowledge transfer,'' in \emph{Proceedings of
  the IEEE Conference on Computer Vision and Pattern Recognition}, 2018, pp.
  9359--9367.

\bibitem{carlucci2019domain}
F.~M. Carlucci, A.~D'Innocente, S.~Bucci, B.~Caputo, and T.~Tommasi, ``Domain
  generalization by solving jigsaw puzzles,'' in \emph{Proceedings of the IEEE
  Conference on Computer Vision and Pattern Recognition}, 2019, pp. 2229--2238.

\bibitem{liu2020weakly}
W.~Liu, C.~Zhang, G.~Lin, T.-Y. Hung, and C.~Miao, ``Weakly supervised
  segmentation with maximum bipartite graph matching,'' in \emph{Proceedings of
  the 28th ACM International Conference on Multimedia}, 2020, pp. 2085--2094.

\bibitem{zhang2020causal}
D.~Zhang, H.~Zhang, J.~Tang, X.~Hua, and Q.~Sun, ``Causal intervention for
  weakly-supervised semantic segmentation,'' in \emph{Advances in Neural
  Information Processing Systems}, 2020.

\bibitem{hariharan2011semantic}
B.~Hariharan, P.~Arbel{\'a}ez, L.~Bourdev, S.~Maji, and J.~Malik, ``Semantic
  contours from inverse detectors,'' in \emph{2011 International Conference on
  Computer Vision}.\hskip 1em plus 0.5em minus 0.4em\relax IEEE, 2011, pp.
  991--998.

\bibitem{lin2014microsoft}
T.-Y. Lin, M.~Maire, S.~Belongie, J.~Hays, P.~Perona, D.~Ramanan,
  P.~Doll{\'a}r, and C.~L. Zitnick, ``Microsoft coco: Common objects in
  context,'' in \emph{European conference on computer vision}.\hskip 1em plus
  0.5em minus 0.4em\relax Springer, 2014, pp. 740--755.

\bibitem{he2016deep}
K.~He, X.~Zhang, S.~Ren, and J.~Sun, ``Deep residual learning for image
  recognition,'' in \emph{Proceedings of the IEEE conference on computer vision
  and pattern recognition}, 2016, pp. 770--778.

\bibitem{deng2009imagenet}
J.~Deng, W.~Dong, R.~Socher, L.-J. Li, K.~Li, and L.~Fei-Fei, ``Imagenet: A
  large-scale hierarchical image database,'' in \emph{2009 IEEE conference on
  computer vision and pattern recognition}.\hskip 1em plus 0.5em minus
  0.4em\relax Ieee, 2009, pp. 248--255.

\bibitem{he2015delving}
K.~He, X.~Zhang, S.~Ren, and J.~Sun, ``Delving deep into rectifiers: Surpassing
  human-level performance on imagenet classification,'' in \emph{Proceedings of
  the IEEE international conference on computer vision}, 2015, pp. 1026--1034.

\bibitem{chang2020mixup}
Y.-T. Chang, Q.~Wang, W.-C. Hung, R.~Piramuthu, Y.-H. Tsai, and M.-H. Yang,
  ``Mixup-cam: Weakly-supervised semantic segmentation via uncertainty
  regularization,'' in \emph{The British Machine Vision Conference}, 2020.

\bibitem{papandreou2015weakly}
G.~Papandreou, L.-C. Chen, K.~P. Murphy, and A.~L. Yuille, ``Weakly-and
  semi-supervised learning of a deep convolutional network for semantic image
  segmentation,'' in \emph{Proceedings of the IEEE international conference on
  computer vision}, 2015, pp. 1742--1750.

\bibitem{ke2021universal}
T.-W. Ke, J.-J. Hwang, and S.~X. Yu, ``Universal weakly supervised segmentation
  by pixel-to-segment contrastive learning,'' in \emph{Proceedings of the
  International Conference on Learning Representations}, 2021.

\bibitem{wang2019boundary}
B.~Wang, G.~Qi, S.~Tang, T.~Zhang, Y.~Wei, L.~Li, and Y.~Zhang, ``Boundary
  perception guidance: a scribble-supervised semantic segmentation approach,''
  in \emph{IJCAI International Joint Conference on Artificial Intelligence},
  2019.

\bibitem{wei2016stc}
Y.~Wei, X.~Liang, Y.~Chen, X.~Shen, M.-M. Cheng, J.~Feng, Y.~Zhao, and S.~Yan,
  ``Stc: A simple to complex framework for weakly-supervised semantic
  segmentation,'' \emph{IEEE transactions on pattern analysis and machine
  intelligence}, vol.~39, no.~11, pp. 2314--2320, 2016.

\bibitem{zhang2020splitting}
T.~Zhang, G.~Lin, W.~Liu, J.~Cai, and A.~Kot, ``Splitting vs. merging: Mining
  object regions with discrepancy and intersection loss for weakly supervised
  semantic segmentation,'' in \emph{European Conference on Computer Vision},
  2020.

\bibitem{fanemploying}
J.~Fan, Z.~Zhang, and T.~Tan, ``Employing multi-estimations for
  weakly-supervised semantic segmentation,'' in \emph{European Conference on
  Computer Vision}, 2020.

\bibitem{wu2019wider}
Z.~Wu, C.~Shen, and A.~Van Den~Hengel, ``Wider or deeper: Revisiting the resnet
  model for visual recognition,'' \emph{Pattern Recognition}, vol.~90, pp.
  119--133, 2019.

\bibitem{simonyan2014very}
K.~Simonyan and A.~Zisserman, ``Very deep convolutional networks for
  large-scale image recognition,'' in \emph{Proceedings of the International
  Conference on Learning Representations}, 2015.

\bibitem{krahenbuhl2011efficient}
P.~Kr{\"a}henb{\"u}hl and V.~Koltun, ``Efficient inference in fully connected
  crfs with gaussian edge potentials,'' in \emph{Advances in Neural Information
  Processing Systems}, 2011, pp. 109--117.

\end{thebibliography}
	
	\begin{IEEEbiography}[{\includegraphics[width=1in,height=1.25in,clip,keepaspectratio]{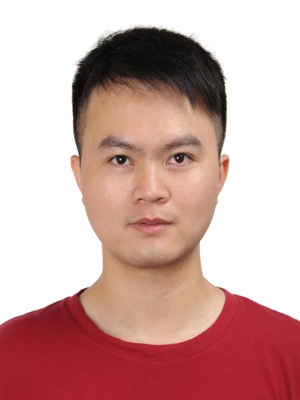}}]{Jinlong Li} received the B.E. degree from the College of Physics and Optoelectronic Engineering, Shenzhen University, China, in 2017. He is currently pursuing the M.S. degree from the College of Computer Science and Software Engineering, Shenzhen University, China. His research interests include weakly-supervised 2D/3D detection/segmentation.
	\end{IEEEbiography}
	\vspace{-0.5in}
	
	\begin{IEEEbiography}[{\includegraphics[width=1in,height=1.25in,clip,keepaspectratio]{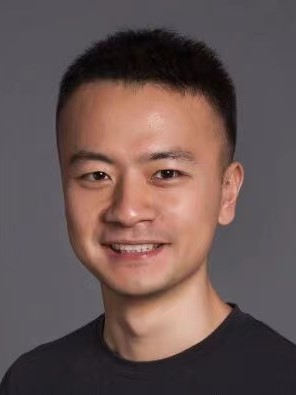}}]{Zequn Jie} received the B.E. degree from the University of Science and Technology of China, Hefei, China, and the Ph.D. degree from the National University of Singapore, Singapore. He was a Post-Doctoral Research Fellow with the Department of Electrical and Computer Engineering, National University of Singapore. He is currently a senior algorithm expert in Meituan Inc. Prior to coming to Meituan, he was a senior researcher in Tencent AI Lab. His research interests mainly fall in the fundamental computer vision topics, e.g. supervised and weakly-supervised object detection, localization and semantic segmentation. He regularly serves as a reviewer of several top-tier conferences and journals, e.g. CVPR, ICCV, ECCV, NeurIPS, ICML, TPAMI.
	\end{IEEEbiography}
	\vspace{-0.5in}
	
	\begin{IEEEbiography}[{\includegraphics[width=1in,height=1.25in,clip,keepaspectratio]{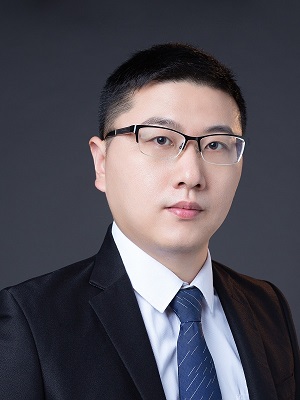}}]{Xu Wang}(M'15) received the B.S. degree from South China Normal University, Guangzhou, China, in 2007, and M.S. degree from Ningbo University, Ningbo, China, in 2010. He received his Ph.D. degree from the Department of Computer Science, City University of Hong Kong in 2014. In 2015, he joined the College of Computer Science and Software Engineering, Shenzhen University, where he is currently an Associate Professor. His research interests are video coding and 3D vision.
	\end{IEEEbiography}
	\vspace{-0.5in}
	
	\begin{IEEEbiography}[{\includegraphics[width=1in,height=1.25in,clip,keepaspectratio]{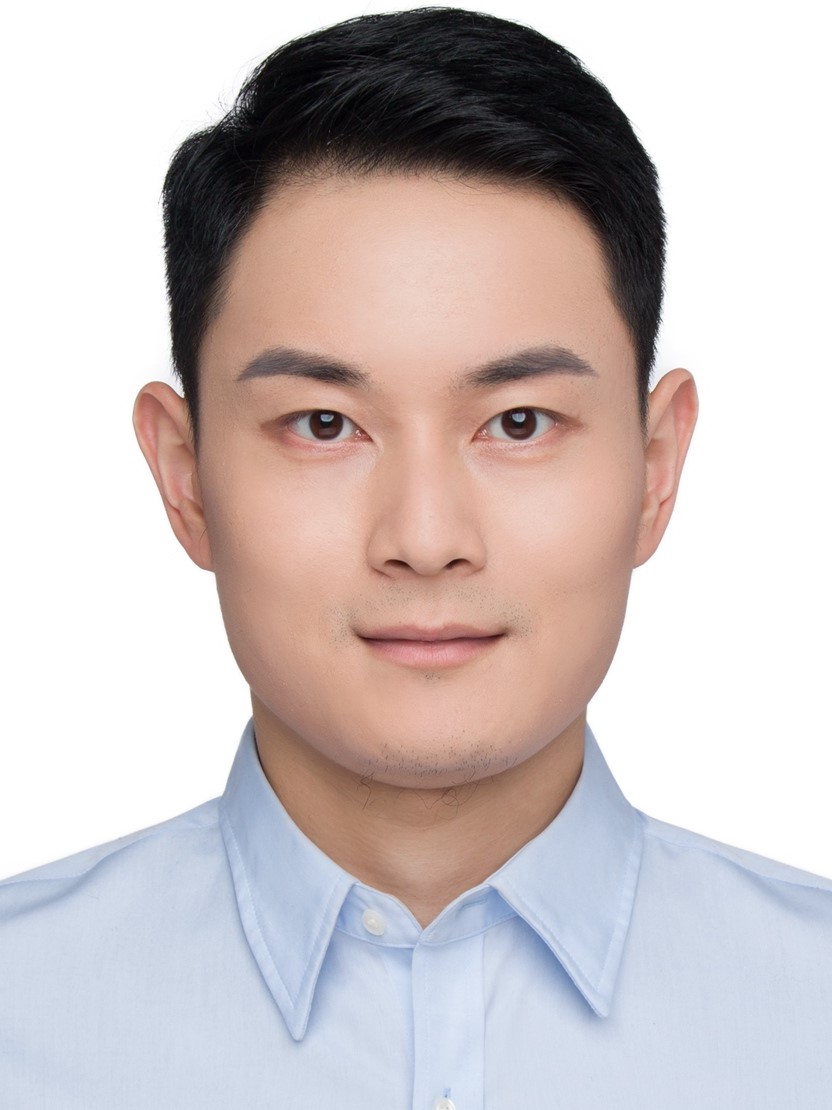}}]{Yu Zhou}(M'19) received the B.S. degree in electronics and information engineering and the M.S. degree in circuits and systems from Xidian University, Xi’an, China, in 2009 and 2012, respectively, and the Ph.D. degree in computer science from the City University of Hong Kong, Hong Kong, in 2017. He is currently an Assistant Professor with Shenzhen University, Shenzhen, China. His current research interests include computational intelligence, machine learning and intelligent information processing.
	\end{IEEEbiography}
	\vspace{-0.5in}
	
	\begin{IEEEbiography}[{\includegraphics[width=1in,height=1.25in,clip,keepaspectratio]{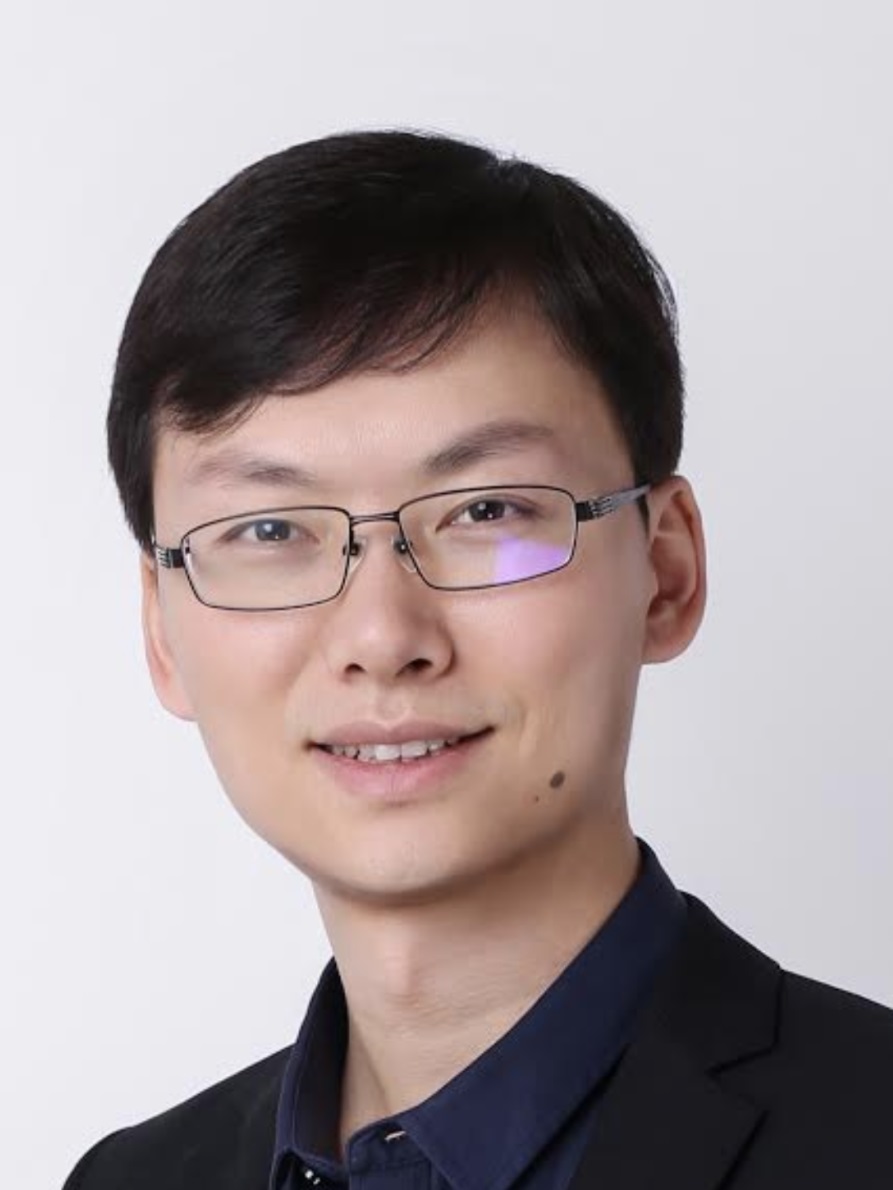}}]{Xiaolin Wei} received Ph.D. degree in Computer Science from Texas A\&M University. His research area includes computer vision, machine learning, computer graphics, 3D vision, augmented reality. He worked as a research engineer at Google, Virtroid and Magic Leap, and now is working at Meituan AI Lab.
	\end{IEEEbiography}
	\vspace{-0.5in}
	
	\begin{IEEEbiography}[{\includegraphics[width=1in,height=1.25in,clip,keepaspectratio]{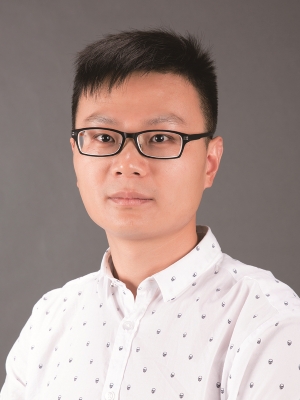}}]{Lin Ma} (M’13)  received the Ph.D. degree from the Department of Electronic Engineering, The Chinese University of Hong Kong, in 2013, the B.E. and M.E. degrees in computer science from the Harbin Institute of Technology, Harbin, China, in 2006 and 2008, respectively. He is now a Researcher with Meituan, Beijing, China.  Previously, he was a Principal Researcher with Tencent AI Laboratory, Shenzhen, China from Sept. 2016 to Jun. 2020. He was a Researcher with the Huawei Noah’Ark Laboratory, Hong Kong, from 2013 to 2016.  His current research interests lie in the areas of computer vision, multimodal deep learning, specifically for image and language, image/video understanding, and quality assessment.
		
		Dr. Ma received the Best Paper Award from the Pacific-Rim Conference on Multimedia in 2008. He was a recipient of the Microsoft Research Asia Fellowship in 2011. He was a finalist in HKIS Young Scientist Award in engineering science in 2012.
	\end{IEEEbiography}
	\vspace{-0.5in}

		
		
	
	%
	
	
	
	
	
	
	

\end{document}